\pdfoutput=1

\documentclass[11pt]{article}

\usepackage[final]{acl}

\usepackage{times}
\usepackage{latexsym}

\usepackage[T1]{fontenc}

\usepackage[utf8]{inputenc}

\usepackage{microtype}

\usepackage{inconsolata}

\usepackage{graphicx}

\usepackage{amsmath}
\usepackage{amssymb}
\usepackage{array}
\usepackage{rotating}
\usepackage[most]{tcolorbox}
\usepackage{colortbl}
\usepackage{soul}
\usepackage{hyperref}

\sethlcolor{gray!20}
\newcolumntype{R}[1]{>{\raggedleft\arraybackslash}p{#1}}

\newcounter{examplecounter}
\newenvironment{example}[1]{\noindent 
$\blacktriangledown$ \textbf{Example \theexamplecounter\stepcounter{examplecounter}:} #1

\small \noindent \\ \rule{0.1\textwidth}{0.5pt} \vspace{1mm} \leftskip3mm \parindent0pt \\}{\par \raggedleft \rule{0.1\textwidth}{0.5pt}\par\vspace{3mm}}

\title{GIER: Gap-Driven Self-Refinement for Large Language Models}

\author{Rinku Dewri \\
  University of Denver, Denver, CO 80208 \\
  \href{mailto:rinku.dewri@du.edu}{rinku.dewri@du.edu}
  }

\begin{document}
\maketitle
\begin{abstract}
We introduce GIER (Gap-driven Iterative Enhancement of Responses), a general framework for improving large language model (LLM) outputs through self-reflection and revision based on conceptual quality criteria. Unlike prompting strategies that rely on demonstrations, examples, or chain-of-thought templates, GIER utilizes natural language descriptions of reasoning gaps, and prompts a model to iteratively critique and refine its own outputs to better satisfy these criteria. Across three reasoning-intensive tasks (SciFact, PrivacyQA, and e-SNLI) and four LLMs (GPT-4.1, GPT-4o Mini, Gemini 1.5 Pro, and Llama 3.3 70B), GIER improves rationale quality, grounding, and reasoning alignment without degrading task accuracy. Our analysis demonstrates that models can not only interpret abstract conceptual gaps but also translate them into concrete reasoning improvements. 
\end{abstract}

\begin{figure*}[t]
\centering
\includegraphics[scale=0.25]{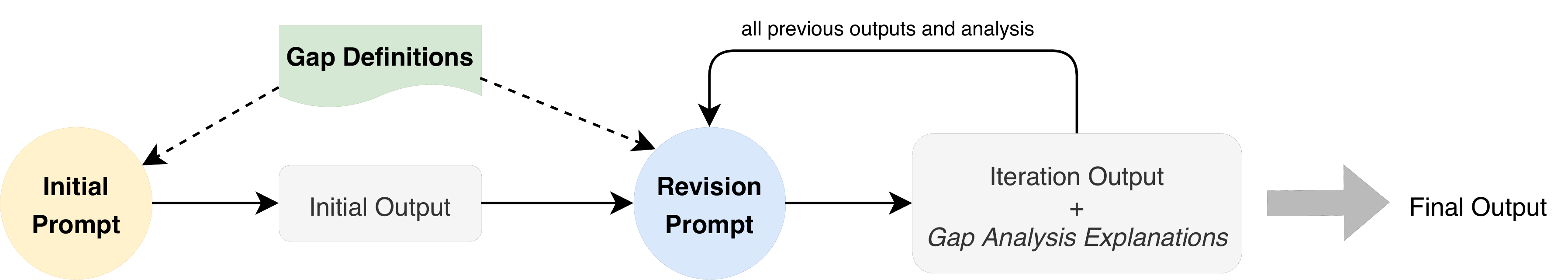}
\caption{Schematic of GIER (Gap-driven Iterative Enhancement of Responses) workflow. An initial prompt is used to obtain an output which is then revised through a revision prompt. Both prompts are provided with conceptual gap definitions. Revisions are repeated over iterations.}
\label{fig:gier_schematic}
\end{figure*}

\section{Introduction}
Recent advances in large language models (LLMs) have significantly improved performance on various natural language processing (NLP) tasks. However, a primary challenge in LLM development is the opacity of their internal reasoning pathways. The challenge lies in the fact that transparent reasoning can have varying expectations based on a task at hand, and may require surfacing evidence in diverse reasoning formats, such as fact-based evidence verification demanding precise textual grounding, domain-specific sentence selection demanding thematic relevance, and common-sense inferential reasoning emphasizing logical coherence and coverage of explanatory steps. 

Recent work has sought to improve LLM reasoning capabilities through various techniques. Chain-of-thought prompting facilitates stepwise reasoning, while self-consistency decoding enhances output robustness by aggregating diverse reasoning paths. Other strategies involve self-refinement, where models critique and revise their responses, or leverage preference-based fine-tuning and supervised training on rationales. These methods typically rely on exemplars or prompt markers to elicit reasoning, or use implicit supervision through preference tuning. However, they do not make explicit the kinds of reasoning flaws a model should avoid.

This raises a central question in LLM interaction paradigms: \textit{Can large language models leverage abstract, natural language descriptions of reasoning gaps to identify and fix flaws in their outputs?}  More specifically, can we guide a model to reason along specific dimensions without relying on examples, demonstrations, or explicit corrective feedback? For example, a prompt to improve the readability of a piece of text may state

\begingroup
\addtolength\leftmargini{-0.2in}
\begin{quote}
    \textit{``To improve readability, break long sentences into shorter ones, organize ideas in a logical sequence, and use transition phrases to connect thoughts smoothly.''}
\end{quote}
\endgroup
\noindent offering explicit revision instructions. In contrast, an abstract gap definition may state
\begingroup
\addtolength\leftmargini{-0.2in}
\begin{quote}
    \textit{``The response lacks clear and coherent flow, making it difficult for readers to follow the reasoning or narrative.''}
\end{quote}
\endgroup
\noindent leaving the identification of flaws and required corrective actions to the model itself. A direct artifact of this self-discovery process would be a reasoning transcript aligned with one or more specific gaps.

With this question in mind, we introduce GIER (Gap-driven Iterative Enhancement of Responses), a model-agnostic prompting framework that improves model outputs via structured self-assessment and revision. Unlike prior work that provides feedback or demonstrations, GIER supplies only natural language descriptions of reasoning gaps, and asks the model to critique and revise its own outputs iteratively. Figure~\ref{fig:gier_schematic} illustrates this workflow. This design probes whether models can internalize and act upon conceptual quality standards.

To demonstrate GIER's versatility, we evaluate it across three diverse reasoning tasks---(i) rationale generation with precise evidence identification and justification for scientific claims (SciFact), (ii) extracting conceptually relevant sentences from a privacy policy based on a question (PrivacyQA), and (iii) explainable natural language inference (NLI) emphasizing the discovery of human-aligned reasoning pathways (e-SNLI). 
Our experiments span four LLMs, including proprietary (GPT-4.1, GPT-4o mini, Gemini 1.5 Pro) and open-weight models (Llama 3.3 70B), highlighting robustness across architectures. In GPT-4.1, compared to a generic prompt, we observe a $63\% \rightarrow 85\%$ improvement in recall of textually grounded rationales in SciFact, a $44\% \rightarrow 66\%$ improvement in extraction of relevant sentences to a PrivacyQA query, with false positives centering on conceptually relevant sentences, and a $62\% \rightarrow 71\%$ alignment of model to human reasoning in e-SNLI natural language inferences. 

To summarize, the core contributions of this work are as follows: (1) we introduce GIER, a framework for improving LLM outputs through self-guided revision based on abstract reasoning gaps, in contrast to exemplar-driven approaches; (2) we design an iterative prompting strategy that elicits structured self-evaluation and revision without fine-tuning or supervision; (3) we define gap definitions for three distinct reasoning tasks and show GIER's effectiveness in aligning model outputs to task-specific reasoning standards; (4) we provide a comprehensive manual evaluation highlighting the importance of soft metrics in assessing explanation fidelity and reasoning quality. 
    
The remainder of the paper is organized in the following manner: Section~\ref{sec:GIER_framework} presents the details of the GIER framework, followed by a comparison with related approaches in Section~\ref{sec:related_work}. In Section~\ref{sec:evaluation_tasks_and_datasets}, we present the three tasks and datasets we use for evaluation, followed by the relevant metrics in Section~\ref{sec:evaluation_metrics}. Section~\ref{sec:results} presents the core results, followed by a discussion in Section~\ref{sec:discussion}. Section~\ref{sec:conclusion} concludes the paper, with a discussion on limitations and future work continued in Section~\ref{sec:limitations}.

\section{The GIER Framework}
\label{sec:GIER_framework}

Iterative self-evaluation in GIER is grounded in externally provided conceptual gap definitions. We define a \textit{conceptual gap} as a natural language description of an abstract standard that responses should ideally satisfy, such as factual accuracy, logical coherence, or answer completeness. These gaps are human-readable, generalizable, and semantically grounded, enabling both LLMs and human evaluators to reason about quality in principled terms. For example, a thematic overreach gap in privacy question-answering may read as: \textit{``The selected sentences cross into privacy-related themes that significantly widen the scope of the provided response.''} 
Appendix~\ref{apdx:gap_definitions} provides the set of gap definitions that we use for the evaluation tasks in this work. 

Given a set of conceptual gap definitions, the GIER process proceeds as follows (Figure~\ref{fig:gier_schematic}).

\noindent \textbf{Initial output:} Following an initial prompt (Appendix~\ref{apdx:prompts}), the LLM generates a response to a given task. The prompt indicates that the gaps should be used as an evaluative context.

A revision prompt then initiates the following steps. The prompt includes all analysis, self-explanations and outputs that the model generated in earlier revisions.

\noindent \textbf{Gap analysis and explanations:} The LLM introspectively scores its most recent response for each gap (0--10), identifying strengths and weaknesses. For each score, the LLM explains its reasoning, referencing specific parts of its response and the relevant gap criterion.

\noindent \textbf{Consolidated explanation:} The LLM writes a consolidated explanation of how it will revise the output based on the gap analysis.

\noindent \textbf{Response revision:} Based on its analysis, the LLM revises its response to better address the gaps, if necessary.

\noindent \textbf{Iterative refinement:} This cycle---analysis, explanation, consolidation, and revision---repeats for a fixed number of rounds or until a stopping criterion is met (e.g., score improvement plateaus).

\section{Related Work}
\label{sec:related_work}

Our work connects to several lines of research on self-refinement, explanation supervision, and alignment. Chain-of-thought prompting \citep{wei2022chain} and its extensions \citep{wang2023self, zelikman2022star, khot2023decomposed} guide models with intermediate reasoning steps or structured outputs but rely on demonstrations or task-specific scaffolds. In contrast, GIER uses only natural language descriptions of reasoning gaps, requiring models to self-diagnose and revise without examples. Self-refinement methods \citep{madaan2023self} show that models can improve through iterative critique or feedback, often using instructive signals or revision exemplars. GIER differs by eliminating both: models must infer flaws and corrective actions directly from abstract quality criteria. Rationale supervision work, such as e-SNLI \citep{camburu2018snli} and ERASER \citep{deyoung2020eraser}, provides gold explanations for post hoc evaluation or training but does not embed quality criteria into the generation process itself. GIER advances this by operationalizing conceptual reasoning gaps as self-assessment tools within the generation loop. Our approach also complements alignment methods like RLHF \citep{ouyang2022training} and preference modeling \citep{rafailov2023dpo}, offering a lightweight alternative that aligns model reasoning through prompts rather than fine-tuning or reward signals. An extended discussion of these related works appear in Appendix~\ref{apdx:related_work}.

\section{Evaluation Tasks and Datasets}
\label{sec:evaluation_tasks_and_datasets}

We evaluate GIER on three tasks: scientific claim verification (using SciFact), privacy-focused question answering (using PrivacyQA), and explainable natural language inference (using e-SNLI). SciFact emphasizes textual grounding and multi-evidence synthesis in scientific claim verification. PrivacyQA probes contextual coverage and thematic relevance in real-world, policy-driven question answering. e-SNLI targets commonsense inference and explanation faithfulness, enabling direct comparison to human-authored justifications. Together, these tasks span abstractive and extractive settings, structured and unstructured evidence, and both objective and subjective reasoning challenges.

\paragraph{Sample Size.} Given the exploratory nature of this work, the high computational cost of iterative LLM experimentation, and the human effort required to validate reasoning alignment, we select a balanced sample of 100 instances per task. While larger-scale evaluations are desirable, prior work \citep{rudin2019stop, wiegreffe2021measuring} has shown that even small samples, when selected for conceptual diversity and difficulty, yield robust insights into model behavior. Our sampling strategies are tailored to each dataset to ensure coverage across label types, reasoning challenges, and task-specific nuances. 

\paragraph{LLM Models.} To support GIER's model-agnostic design, we evaluate across four large language models: (i) OpenAI's GPT-4.1, a state-of-the-art commercial model; (ii) OpenAI's GPT-4o mini, a more cost-effective baseline; (iii) Google's Gemini 1.5 Pro, a smaller model with differing architecture (mixture of experts); and (iv) Meta's Llama 3.3 70B, a widely adopted open-source model for reasoning. Discussions are primarily based on the GPT-4.1 model, unless stated otherwise. All models are accessed via a Python API, with a temperature setting of zero.

\paragraph{Gap Identification.} To determine which conceptual gaps are most relevant for each task, we conducted a pilot study using 25 randomly sampled instances from the datasets per task. We ran GIER using ChatGPT via its web interface and manually examined the effectiveness of an initial set of candidate gaps. Based on these findings, we iteratively refined the gap definitions to ensure conceptual clarity and practical utility. Reflections on this gap selection process are provided in Appendix~\ref{apdx:reflections_gap_writing}. 

\subsection{SciFact: Scientific Rationale Generation}
Each instance in the SciFact dataset \citep{wadden2020scifact} comprises a scientific abstract, a claim, and a gold label indicating whether the abstract supports or refutes the claim. Gold rationales consist of sets of one or more sentences from the abstract, each signifying an independent evidence line. The LLM is tasked with predicting the correct label and generating a rationale to justify its decision.

We sample 100 instances, ensuring a balanced label distribution (50 support, 50 refute) and a diversity of gold rationale complexity: 25 instances with a single evidence rationale set, 29 with two, and 46 with three or more. 
We use five conceptual gaps in this task: \textit{coverage}, \textit{conciseness}, \textit{textual grounding}, \textit{source faithfulness}, and \textit{unsupported emphasis}. Full definitions of these gaps are provided in Appendix~\ref{apdx:gap_definitions_scifact}. Textual grounding (quoting text from the abstract) is essential in this task for automatic overlap evaluation. Source faithfulness is critical in scientific domains, while the other gaps ensure the rationale is focused, balanced, and justified.

\subsection{PrivacyQA: Extractive Sentence Selection}
PrivacyQA is a question-answering dataset focused on privacy policies, containing crowd-sourced questions, an answerability label, and if answerable, sentence-level gold answers extracted from corresponding policy documents \citep{ravichander2019privacyqa}. Each question is tagged with one or more thematic categories, such as First-Party Collection/Use (FPCU), Third-Party Sharing/Collection (TPSC), Data Security (DS), Data Retention (DR), User Choice/Control (UCC), and User Access/Edit/Deletion (UAED). 
The LLM's task is to assign an answerability label and, if answerable, select a subset of sentences from the policy that best answers a given question. 

We cluster samples based on the question's theme and answerability, and then sample evenly across all clusters, resulting in 100 instances. Since some questions span multiple themes, our final selection includes 26 FPCU questions, 22 TPSC, 21 DS, 24 DR, 21 UCC, and 18 UAED questions, covering all 27 privacy policies. We identified a few cases where the gold answers omitted semantically relevant sentences (e.g., a question about DS contained only cookie reference sentences). These were manually corrected by adding, but not deleting, theme-aligned sentences in the gold answer set prior to evaluation. 

We apply two conceptual gaps to this task: \textit{coverage} and \textit{thematic overreach}. These gaps are chosen to reflect a real-world tension in privacy Q\&A, where models must infer sufficient context without overstepping the thematic boundaries defined by user expectations. As such, coverage promotes inclusion of semantically relevant context, while thematic overreach penalizes inclusion of sentences that stray beyond the question's core theme. Full descriptions of the gaps are in Appendix~\ref{apdx:gap_definitions_privacyqa}.

\subsection{e-SNLI: Explainable Commonsense Inferences}
e-SNLI \citep{camburu2018snli} is an extension of SNLI \citep{bowman2015snli}, where each instance includes a premise, a hypothesis, a gold label (entailment, contradiction, or neutral), and a free-text human rationale explaining the label. 
Unlike standard NLI, our focus is on assessing whether the LLM's reasoning mirrors that of humans. The LLM is tasked with providing an entailment score between 0 and 10---$[0,1]$=entailment, $[9,10]$=contradiction, otherwise neutral---and a reason for the score.

We begin by randomly sampling 1,000 instances, stratified across the three labels. Using GPT-4.1, we run a standard NLI task to determine which instances it can or cannot label correctly. 827 are classified correctly, suggesting that many examples may be ``easy'' cases for the LLM. We focus our evaluation on harder instances, selecting 75 failure cases and 25 success cases, uniformly distributed across the three label types. 

To define conceptual gaps for this task, we draw on the inference ontology developed by \citet{nie2020adversarial} in the ANLI dataset. This results in five gap types, capturing diverse reasoning pathways: \textit{quantitative and comparative reasoning}, \textit{logical inferences}, \textit{lexical inferences}, \textit{pragmatic inferences}, and \textit{reference resolution}. Full gap definitions are provided in Appendix~\ref{apdx:gap_definitions_esnli}.

\section{Evaluation Metrics}
\label{sec:evaluation_metrics}

We perform both quantitative and qualitative evaluations of GIER across the three tasks. Metrics are defined per task to effectively measure improvements in reasoning alignment and addressing conceptual gaps.

\paragraph{Baselines.} For all tasks, the baselines for improvement are the outputs using an initial prompt, with or without the gap definitions. Subsequent improvements are measured relative to these baselines. 

\paragraph{Iterations.} We initially run GIER's iterative process for five iterations after the initial prompt. We did not attempt more iterations since results at this stage indicated diminishing returns.

\subsection{SciFact Rationale Generation}

We evaluate GIER's impact on scientific rationale quality using measures aligned with one or more targeted reasoning gaps. 

\noindent \textbf{Decision Accuracy:} Proportion of samples with correctly predicted label, indicating alignment with gold annotations.

\noindent \textbf{Rationales Recall} (Coverage): Proportion of gold rationales (independent evidence sets) where all sentences in a set are at least partially quoted in the model's rationale, capturing full recovery of each reasoning path.

\noindent \textbf{Grounding Ratio} (Textual Grounding, Conciseness): Proportion of model rationale tokens that match spans in the abstract. Higher values reflect stronger grounding and penalize over-elaboration.

In addition, we manually analyze the model rationales to evaluate if outputs for multi-gold rationale samples are simply enumerations or includes logical bridges. 

\subsection{PrivacyQA Sentence Selection}

When evaluating GIER's performance on privacy question answering, we emphasize performance that is more thematic and user-focused, leading to a more domain-specific assessment. 

\noindent \textbf{Decision Accuracy:} Proportion of questions with correct prediction on whether it is answerable from the policy text.

\noindent \textbf{Selection Precision and Recall} (Coverage): Standard metrics evaluating the correctness and completeness of the selected sentences in answerable questions.

\noindent \textbf{Thematic Drift} (Thematic Overreach): Quantifies how far false positive answer sentences stray from the thematic type of the question, indicating overreach. A value of zero implies complete alignment; $\approx 1$ implies sentences are in the conceptual neighborhood of the question. Appendix~\ref{apdx:thematic_drift} discusses the metric computation in depth.

To supplement automated metrics, we also conduct a manual evaluation of the unanswerable samples for which a model produces an answer (false positives). The goal is to check if answers are truly irrelevant or reflect alternative interpretations of questions. 

\begin{table*}[t]
\begin{centering}
\begin{tabular}{rcccc}
\hline 
\multicolumn{5}{c}{\textbf{SciFact rationale generation}}\tabularnewline
\hline 
 & \textit{GPT-4.1} & \textit{GPT-4o Mini} & \textit{Gemini 1.5 Pro} & \textit{Llama 3.3 70B}\tabularnewline
Decision Accuracy & $0.91\rightarrow0.92\rightarrow0.92$ & $0.87\rightarrow0.87$ & $0.92\rightarrow0.91$ & $0.91\rightarrow0.91$ \tabularnewline
Rationales Recall & $0.63\rightarrow\mathbf{0.74\rightarrow0.85}$ & $0.52\rightarrow0.59$ & $0.63\rightarrow0.68$ & $0.57\rightarrow0.64$ \tabularnewline
Grounding Ratio & $0.54\rightarrow0.58\rightarrow0.63$ & $0.48\rightarrow0.36$ & $0.43\rightarrow0.41$ & $0.49\rightarrow0.38$ \tabularnewline
\hline 
\multicolumn{5}{c}{\textbf{PrivacyQA sentence selection}}\tabularnewline
\hline 
 & \textit{GPT-4.1} & \textit{GPT-4o Mini} & \textit{Gemini 1.5 Pro} & \textit{Llama 3.3 70B}\tabularnewline
Decision Accuracy & $0.70\rightarrow0.70\rightarrow0.63$ & $0.61\rightarrow0.59$ & $0.65\rightarrow0.64$ & $0.61 \rightarrow 0.53$ \tabularnewline
Selection Precision & $0.68\rightarrow0.68\rightarrow0.53$ & $0.76\rightarrow0.57$ & $0.69\rightarrow0.52$ & $0.61 \rightarrow 0.51$\tabularnewline
Selection Recall & $0.44\rightarrow\mathbf{0.45\rightarrow0.66}$ & $0.38\rightarrow0.45$ & $0.48\rightarrow0.56$ & $0.40 \rightarrow 0.43$\tabularnewline
Thematic Drift & $0.67\rightarrow0.69\rightarrow1.36$ & $0.41\rightarrow1.45$ & $0.69\rightarrow1.21$ & $0.78 \rightarrow 1.21$\tabularnewline
\hline 
\multicolumn{5}{c}{\textbf{e-SNLI commonsense inference}}\tabularnewline
\hline 
 & \textit{GPT-4.1} & \textit{GPT-4o Mini} & \textit{Gemini 1.5 Pro} & \textit{Llama 3.3 70B}\tabularnewline
Decision Accuracy & $0.57\rightarrow0.57\rightarrow0.55$ & $0.60\rightarrow0.52$ & $0.56\rightarrow0.57$ & $0.54\rightarrow0.61$\tabularnewline
Reasoning Attribution & $\mathbf{0.62\rightarrow0.75}\rightarrow0.71$ & $0.51\rightarrow0.62$ & $0.63\rightarrow0.73$ & $0.50\rightarrow0.75$\tabularnewline
Best Alignment & $0.12\ |\ 0.47\ |\ 0.30$ & $0.21\ |\ 0.51$ & $0.40\ |\ 0.39$ & $0.18\ |\ 0.59$\tabularnewline
\hline 
\end{tabular}
\par\end{centering}
\caption{Quantitative metrics for three GIER tasks for different LLM models.
The three values for GPT-4.1 represents the baseline without gap definitions,
then baseline with gap definitions, and the final GIER iteration.
Values for other models represent the baseline with gap definitions
and the final GIER iteration.}

\label{tab:result_summary}
\end{table*}

\subsection{e-SNLI Commonsense Inference}

For this task, we abstract to a meta-level evaluation of overall reasoning alignment, assessing how well a model's reasoning emulates human provided reasoning patterns. 

\noindent \textbf{Decision Accuracy:} Proportion of samples with correctly predicted NLI label compared to the gold label.

\noindent \textbf{Reasoning Attribution:} Using the GPT-4.1 LLM, we segment each human rationale, LLM rationale, and the LLM's gap-level explanations into reasoning chunks. For example, the statements \textit{``Young boy refers to kids. Kicking a soccer implies kids playing soccer.''} are broken into two reasoning chunks---\textit{``Young boy refers to kids.''} and \textit{``Kicking a soccer implies kids playing soccer.''} Then, for a given sample, we use the \textsf{
DeBERTa-v3-large-mnli-fever-anli-ling-wanli} model to obtain the NLI scores considering an LLM chunk as the premise and a human chunk as the hypothesis. The human reasoning chunk is attributed to the LLM chunk if there is strong entailment or contradiction (threshold > $80\%$), reflecting the LLM’s capacity to surface the human reasoning step, whether in agreement or disagreement. We measure reasoning attribution rate at both the overall sample level and within each conceptual gap category, enabling insight into which gaps drive alignment. 

\noindent \textbf{Best Alignment:} Proportion of samples per explanation source---baseline without gap definitions, initial GIER iteration, or final GIER iteration--—where a sample achieves the highest attribution (average entailment/contradiction score over the human reasoning chunks). 

We also performed a manual analysis of the reasoning provided for all samples that were incorrectly labeled to see if label mismatch implies faulty reasoning. 

\section{Results}
\label{sec:results}

Table~\ref{tab:result_summary} summarizes the quantitative performance of GIER for the three tasks and across different models. Appendix~\ref{apdx:qualitative_examples} provides representative examples of GIER output on the tasks.

\paragraph{SciFact.} We observe clear benefits from both gap-informed prompting and iterative enhancement. Introducing conceptual gap definitions into the initial prompt increases rationale coverage from $63\%$ to $74\%$. Iterative GIER refinements then lead to further improvement, particularly in rationales recall ($+14.9\%$ gain) and grounding ratio ($+8.6\%$), indicating stronger textual anchoring and conciseness, while decision accuracy remains stable. These results suggest that conceptual scaffolding alone improves reasoning, and that structured revisiting of gaps enables further alignment with both evidence and interpretive structure.

\paragraph{PrivacyQA.} In PrivacyQA, we observe trade-offs between answer completeness and thematic precision. Switching to gap-informed prompting maintains decision accuracy and selection recall, while iterative GIER refinements substantially boost the selection of correct sentences $(+46.7\%$ gain). Precision/recall trade-off is evident in the selection; however, a thematic drift score near 1 indicates that many false positives remain within the conceptual neighborhood of the question, suggesting that precision drops stem from related evidence selection rather than irrelevant noise. Also, the accuracy drop  ($70\% \rightarrow 63\%$) is not due to failures in identifying truly answerable questions, which rises from $41\rightarrow46$ of the 50 answerable questions. Rather, it is due to a rise in cases were unanswerable questions are marked as answerable ($21 \rightarrow 33$ cases). This shift toward more inclusive answering reflects the inherent subjectivity and ambiguity in interpreting privacy policies. 

\paragraph{e-SNLI.} In e-SNLI, decision accuracy remains stable, but human-aligned reasoning discovery improves markedly with GIER. Introducing conceptual gap definitions to the prompt yields a $+20.9\%$ relative gain ($62\%\rightarrow75\%$) in capturing human reasoning chunks, with the final iteration sustaining this improvement. Best alignment further shows that, while $47\%$ of samples achieve their most human-aligned reasoning in the initial gap-informed prompt, $30\%$ peak in the final GIER iteration, indicating the recovery of latent reasoning quality that single-shot prompting could not elicit. 

We note that, although these gains mirror those of prior refinement methods, GIER achieves them without exemplars or pattern-based corrections. This positions GIER as a criterion-driven alternative to methods that rely on demonstration-based or pattern-learning refinements.

\paragraph{Statistical Significance.} 
A Wilcoxon signed-rank test on paired per-example metric values reveals that, across all tasks, variations in decision accuracy between GIER’s initial and final iterations are not statistically significant ($p > 0.10$), indicating that predictions remain stable throughout revisions. In contrast, all recall-oriented metrics show statistically significant improvements ($p < 0.001$), with confidence intervals tightly bounded away from zero. The minor decline in reasoning attribution on e-SNLI is not significant, while the decline in selection precision on PrivacyQA is---though the corresponding confidence interval for mean change in thematic drift ($[+0.33, +1.09]$) suggests that spurious sentences with multiple inferential steps beyond the question are rarely introduced. Full per-task statistics, including effect sizes and confidence bounds, are provided in Appendix~\ref{apdx:statistical_significance}.

\paragraph{Iteration Variations.} 
Across all tasks, we observe that most metrics stabilize by the third iteration, with only minor and statistically insignificant variations thereafter. Tables~\ref{tab:iterations} and \ref{tab:iterations_sig_test} in Appendix~\ref{apdx:iteration_variations} report full metric trajectories and Wilcoxon signed-rank significance tests. In the first iteration, rationale recall increases by up to $+8.3\%$ in SciFact and sentence recall by $+16.6\%$ in PrivacyQA, with moderate gains continuing through the third iteration. In PrivacyQA, the drop in sentence selection precision remains statistically significant ($p<0.05$) up to iteration three, while the increase in thematic drift is only significant in the first iteration. Other metrics, including decision accuracy and e-SNLI reasoning attribution, exhibit no statistically significant changes beyond the initial GIER prompt. 

\paragraph{Model Variations.} 
GIER's effectiveness across models reflects an interplay between model capabilities and task demands. GPT-4.1, with strong reasoning and instruction-following skills, benefits most in SciFact, showing a $+14.9\%$ gain in rationale recall and $+8.6\%$ in grounding ratio. In contrast, Gemini and Llama show smaller recall gains ($+7.9\%$, $+12.2\%$) and reductions in grounding, suggesting a tendency to paraphrase rather than quote directly, possibly due to weaker instruction tuning or a preference for coherence over fidelity. In PrivacyQA, GIER improves recall across models but often at the cost of precision and thematic stability. GPT-4.1 achieves the highest recall gain, though with broader sentence inclusion, indicating a strong response to abstract prompts that sometimes leads to over-selection in ambiguous policy contexts. Llama’s smaller recall gain ($+7.5\%$) and milder drift suggest a more conservative retrieval style. On the reasoning-centric e-SNLI task, Llama improves most in attribution quality ($+50.0\%$), followed by GPT-4o Mini ($+21.5\%$), both from lower baselines. This suggests GIER is especially effective at eliciting justifications when baseline reasoning is weaker. These patterns indicate that, while GIER's model-agnostic structure enables performance gains across diverse models, including substantial scaffolding for weaker systems, it proves most beneficial when a model's strengths align with the reasoning style and fidelity needs of the task. 

\paragraph{Ablation.}
We perform ablation experiments removing individual gap definitions and the explicit gap-based explanation step across the three tasks. Dropping the coverage gap (in SciFact and PrivacyQA) notably reduces retrieval breadth but sometimes improves decision accuracy by encouraging more conservative outputs (PrivacyQA). The conciseness gap in SciFact controls verbosity and grounding, while thematic overreach acts as a soft constraint to suppress conceptual drift in PrivacyQA. The explanation step during revisions consistently emerges as a key driver for identifying multiple reasoning pathways across all tasks. These results underscore that GIER's gap-driven scaffolding not only improves explanation quality but also actively shapes the model's reasoning and decision-making behavior. Table~\ref{tab:ablation_drop} provides the full quantitative results, and a detailed analysis is discussed in Appendix~\ref{apdx:ablation}.

\section{Discussion}
\label{sec:discussion}

\paragraph{SciFact Rationale Connectivity.}
SciFact claims are often justified by multiple gold rationales (evidence sets). We analyzed GPT-4.1's rationales for the correctly labeled samples in which three or more sentences from the abstract contributed to the combined gold rationale. There were 45 such cases. 
In 24 of the 45 samples, the rationale simply enumerates sentences using surface-level transitions such as \textit{``The text states...,''} \textit{``Additionally...,''} or \textit{``The study notes....''} In contrast, the remaining 21 rationales display connective reasoning---linking study steps, highlighting statistical significance, combining fragments from different sentences, adding adverbial clauses of reason (e.g., \textit{``Since the primary outcome did not change,...''}), or inferential cues (e.g., \textit{``This indicates that...''}). These logical bridges primarily emerged during revision and appear to reflect sensitivity to the coverage gap. 
Example 5 in Appendix~\ref{apdx:qualitative_examples} illustrates this kind of reasoning.

\begin{table}[t]
\begin{centering}
\setlength{\tabcolsep}{2.7pt}
\begin{tabular}{p{0.1\textwidth}cccccc}
\hline 
\textbf{Selection Type} & \multicolumn{6}{c}{\textbf{Question Type}}\tabularnewline
\hline 
 & FPCU & TPSC & UCC & UAED & DS & DR\tabularnewline
IG & 8 & 4 & 4 & 4 & 8 & 7\tabularnewline
FPCU & \cellcolor{cyan!50}\textbf{62} & \cellcolor{gray!50}48 & \cellcolor{gray!50}59 & \cellcolor{gray!50}42 & \cellcolor{gray!25}21 & \cellcolor{gray!25}28\tabularnewline
TPSC & 5 & \cellcolor{cyan!50}\textbf{76} & \cellcolor{gray!50}65 & 2 & \cellcolor{gray!25}16 & 10\tabularnewline
UCC & 6 & \cellcolor{gray!25}21 & \cellcolor{cyan!50}\textbf{34} & \cellcolor{gray!25}22 & 9 & 7\tabularnewline
UAED & 0 & 1 & 7 & \cellcolor{cyan!50}\textbf{7} & 3 & 3\tabularnewline
DS & 1 & 8 & 1 & 0 & \cellcolor{cyan!50}\textbf{62} & 11\tabularnewline
DR & 0 & 0 & 2 & 4 & 9 & \cellcolor{cyan!50}\textbf{14}\tabularnewline
ISA & 0 & 4 & 5 & 6 & 6 & 5\tabularnewline
PC & 1 & 0 & 1 & 0 & 2 & 0\tabularnewline
PCI & 0 & 0 & 0 & 3 & 3 & 0\tabularnewline
OTH & 1 & 11 & 2 & 0 & 5 & 0\tabularnewline
\hline 
\end{tabular}
\par\end{centering}
\caption{Selected sentences' type by question type in the final GIER iteration across answered PrivacyQA samples. Highlighted cells reflect significant co-referred privacy themes in the answer for a given question type. Additional abbreviations: IG (Introductory/Generic), ISA (International/Specific Audience, PC (Policy Change), PCI (Privacy Contact Information), and OTH (Others).}
\label{tab:drift_by_type}
\end{table}

\paragraph{PrivacyQA Drift Characterization.}
Table~\ref{tab:drift_by_type} shows the distribution of themes in sentences selected by GPT-4.1 (GIER final iteration) for specific question types. We observe strong thematic alignment for questions about FPCU ($82.6\%$ alignment rate), and TPSC questions show a high concentration of on-theme selections, with frequent but conceptually adjacent spillovers into FPCU and UCC. In contrast, UCC, UAED, and DR questions often yield more diffused answers, with substantial overlap from TPSC and/or FPCU sentences. Privacy policies are relatively scarce in UCC/UAED/DR sentences; the observed drift may therefore reflect not just limitations in model selection, but also structural constraints in the source material. Selection of thematically unrelated sentences such as ISA, PC, PCI, and others are rare, indicating that the outputs remain anchored in relevant zones of privacy discourse. 

A particularly salient pattern is the ubiquitous presence of FPCU sentences across all question types. This reveals a systematic preference for FPCU sentences, likely driven by their abundance in policies (collection/use disclosures are the most verbose), general applicability across many privacy concerns, and perhaps a bias toward safe, foundational disclosures.
Other broader patterns include strong co-selection between TPSC, FPCU, and UCC sentences. This trend reflects the document-level practice of grouping data sharing with collection procedures and user opt-out options. Similarly, DS questions frequently draw from TPSC and FPCU content, reflecting the reality that security guarantees are often embedded in broader discussions about data handling. 

\paragraph{e-SNLI Decision vs. Reasoning Accuracy.}
Despite GPT-4.1 achieving high attribution scores in e-SNLI, it produced incorrect labels for 45 samples. We manually analyzed these cases, categorizing them as \verb+aligned+ (reasoning matches gold, but label differs), \verb+counter+ (model identifies but challenges gold reasoning), or \verb+misaligned+ (reasoning unrelated to gold). 20 cases were \verb+aligned+, e.g., the gold rationale \textit{``Not all beaches are sandy''} supports a neutral label, while the model chose entailment, stating \textit{``The context says the dog is on a beach, but does not specify if it is sandy; most beaches are sandy, so the entailment is very strong but not certain.''} 15 were \verb+counter+, e.g., the model refutes the gold rationale \textit{``Partially-drunk beverages suggests one is sitting at a table drunk''} (entailment) with \textit{``The context only shows young people with partially-consumed drinks, not that they are drunk or that the drinks are alcoholic; the statement claims more than is supported''} (neutral). These cases highlight that incorrect decisions can still reflect grounded reasoning, underscoring the need for evaluation frameworks that go beyond label agreement to assess reasoning quality.

Further discussion items are provided in Appendix~\ref{apdx:extended_discussion}.

\section{Conclusion}
\label{sec:conclusion}

In this work, we presented GIER, a general and model-agnostic framework for improving LLM outputs through iterative self-revision driven by conceptual gap definitions. 
Across three diverse tasks---scientific rationale generation, extractive question answering, and commonsense inference---we showed that GIER consistently improves rationale recall, sentence recall, and alignment with human reasoning. 
While GIER incurs additional computational cost and relies on well-specified gap definitions, it offers a scalable, training-free pathway for enhancing explanation fidelity. Future directions include extending GIER to dialogue, multi-turn reasoning, code generation, and decision-making under uncertainty, as well as developing automated methods for gap prioritization, iteration stopping, and verification. GIER contributes to the broader goal of aligning language model reasoning with human-centric standards of quality, interpretability, and reliability.

\section{Limitations}
\label{sec:limitations}

While the GIER framework demonstrates a promising alternative to example-driven self-correction, several limitations warrant discussion regarding its design, scope, and generalizability.

\paragraph{Interpretability of Conceptual Gaps.} 
GIER's effectiveness depends critically on the clarity, precision, and completeness of the human-authored gap definitions. Although this approach avoids reliance on example-based supervision, it inherits the broader challenges of prompt sensitivity in large language models. The quality of revisions is tightly coupled to how well the conceptual gaps are operationalized in natural language. Applying GIER to new domains or more subtle tasks may require careful prompt engineering or iterative refinement of gap definitions.
        
\paragraph{Dependence on Model Competence.}
GIER operates by guiding models to reflect on and refine their own outputs using internalized knowledge and reasoning patterns. It does not introduce new external information. Consequently, its effectiveness is bounded by the model’s underlying capabilities. In tasks where the model lacks domain knowledge or exhibits poor reasoning skills, GIER may yield limited gains or fail to close critical performance gaps.

\paragraph{Computational Overhead.} GIER relies on an iterative critique-and-revise loop, which introduces additional computational costs relative to single-pass prompting. While manageable in this study, the cost may become prohibitive for large-scale deployments, latency-sensitive applications, or when operating with more expensive frontier models. Strategies for pruning unnecessary iterations or optimizing reflection steps could help mitigate this overhead. Future work should explore dynamic stopping criteria, prompt efficiency improvements, or hybrid approaches combining gap-driven refinement with offline model training to balance performance and cost.

\paragraph{Evaluation Scope.} This study primarily evaluates GIER using models with relatively strong reasoning capabilities. The extent to which the method generalizes to smaller, resource-constrained models remains unexplored. Future work should assess whether GIER can produce reasoning improvements in models with weaker innate capabilities and whether the gap-driven approach remains effective when applied to models further from the frontier.

\paragraph{Thematic Drift.}
Our thematic drift metric, used in the PrivacyQA experiments, is a task-specific heuristic informed by domain expertise. While it captures broad patterns of semantic alignment, it simplifies the complex ways that meaning evolves in privacy policies. Its applicability beyond this domain, and its sensitivity to edge cases or subtle semantic shifts, remains a limitation. Developing more generalizable or model-based measures of thematic coherence is an open avenue.

\paragraph{Position Relative to CoT and Implicit Self-Correction.}
While GIER differs from few-shot or zero-shot chain-of-thought prompting that provides explicit, conceptually grounded feedback, the boundary between explicit gap-driven self-reflection and implicit self-correction via CoT is not always sharply defined. One may argue that advanced CoT or scratchpad prompting can serve similar functions to GIER’s reflection mechanism. Future work could further clarify this distinction through more direct comparisons, both methodologically and empirically.
    
\paragraph{Hallucinating Gap Satisfaction.}
Because gap definitions are abstract and do not inherently enforce factual correctness, models may sometimes generate superficially plausible but incorrect rationales that appear to satisfy the gaps. This risk of hallucinated compliance---where the model optimizes for perceived satisfaction rather than factual soundness---can lead to confident but inaccurate explanations. Gaps like textual grounding and source faithfulness are designed to mitigate this risk by anchoring reasoning to evidence. Nonetheless, reducing error propagation across iterations remains an important challenge that may require complementary safeguards such as verification models or post-hoc audits.

\paragraph{Lack of Gap Prioritization.}
GIER does not impose an explicit hierarchy or prioritization among gaps; the model must navigate trade-offs autonomously during revision. While this design mirrors human-like reasoning under constraint, it also means that the model's preferences are not explicitly steered when gaps are in tension. Our results suggest that models can often improve across multiple gaps simultaneously, but optimal conflict resolution remains an open question. Future extensions of GIER could incorporate dynamic weighting or task-conditioned prioritization of gaps to better align with varying application goals.

\paragraph{Sample Size Constraints.}
We use 100 samples per task, reflecting the high computational cost of multi-step iterative revision. While this sample size limits statistical power, we balance it with in-depth qualitative analysis, multiple reference-based evaluations, and task diversity. Nonetheless, larger-scale evaluations would strengthen the generalizability of our findings and allow for more robust statistical conclusions. This remains an important direction for future validation.

\paragraph{Design Effort Trade-Off.}
While GIER reduces dependence on labeled examples, it introduces an upfront design cost in specifying clear, effective gap definitions for each task. Crafting high-quality conceptual scaffolds requires both domain understanding and careful prompt engineering to ensure that the gaps meaningfully guide revision. This shifts the supervision burden from data annotation to conceptual design. Whether this trade-off is preferable depends on the application context—particularly whether domain expertise is more accessible than large-scale labeled datasets.

\paragraph{Iteration Termination Criteria.}
GIER assumes a fixed number of critique-revision iterations, but does not yet include an adaptive mechanism for determining when revision is sufficient. In some cases, unnecessary iterations may produce diminishing returns, redundant changes, or even slight regressions. Conversely, terminating too early may leave residual gaps unaddressed. Developing reliable stopping criteria---whether through self-assessment, external validation, or heuristic thresholds---is an open direction for improving GIER’s efficiency and robustness in practical settings.

\bibliography{references}

\appendix

\section*{List of Appendices}

\begin{itemize}
    \item \textbf{Appendix~\ref{apdx:related_work}}: Extended Related Work
    \item \textbf{Appendix~\ref{apdx:gap_definitions}}: Gap Definitions
    \item \textbf{Appendix~\ref{apdx:thematic_drift}}: Privacy Thematic Drift
    \item \textbf{Appendix~\ref{apdx:prompts}}: Prompts
    \item \textbf{Appendix~\ref{apdx:statistical_significance}}: Statistical Significance
    \item \textbf{Appendix~\ref{apdx:iteration_variations}}: Iteration Variations
    \item \textbf{Appendix~\ref{apdx:ablation}}: Ablation Study
    \item \textbf{Appendix~\ref{apdx:reflections_gap_writing}}: Reflections on Gap Writing
    \item \textbf{Appendix~\ref{apdx:extended_discussion}}: Extended Discussion
    \item \textbf{Appendix~\ref{apdx:qualitative_examples}}: Qualitative Examples\\
    
\end{itemize}

\section{Extended Related Work}
\label{apdx:related_work}

A broad body of work has sought to improve and evaluate the reasoning abilities of large language models, spanning dimensions such as rationale generation, explanation quality, self-improvement, and alignment with human expectations. We organize prior research into five intersecting themes: quality-centered evaluation, prompting and reasoning scaffolds, natural language feedback and revision, explanation-based supervision, and fine-tuning with human feedback.

\paragraph{Quality-Centered Evaluation.}
Several studies have proposed fine-grained evaluation dimensions such as factual consistency, completeness, relevance, and helpfulness to assess LLM outputs across tasks like summarization and QA \citep{askell2021helpfulness, durmus2020feqa, fabbri2022qafacteval, tam2023evaluating, wang2024evaluating, zhang2024relevance}. These criteria are increasingly adopted in scoring rubrics and human evaluation protocols to go beyond surface-level metrics like BLEU or ROUGE. 
Unlike prior work that typically employs these criteria for scoring alone, GIER integrates them directly into the generation loop, enabling models to self-evaluate and revise based on these abstractions.

\paragraph{Prompting and Reasoning Scaffolds.}
Prompting strategies have been central to eliciting reasoning from LLMs. Chain-of-thought prompting \citep{wei2022chain} encourages step-by-step reasoning, while self-consistency decoding \citep{wang2023self} aggregates diverse reasoning paths to improve robustness. Several extensions have explored structured prompting formats, intermediate representations, or modular reasoning \citep{zelikman2022star, khot2023decomposed, creswell2023selection}. These approaches rely on examples, reasoning templates, or explicit intermediate supervision to guide the model. In contrast, GIER avoids all demonstrations or output examples. It relies instead on general natural language descriptions of reasoning flaws, testing whether LLMs can autonomously diagnose and revise based on high-level quality expectations.

\paragraph{Natural Language Feedback and Iterative Self-Revision.}
Recent work has shown that LLMs can improve their outputs when provided with natural language feedback. \citet{madaan2023self} and \citet{rinaldi2024self} use self-refinement techniques where the model critiques and revises its own answers. They demonstrate performance gains through iterative correction mechanisms, often using structured prompts or meta-feedback. Closely related to our setting, \citet{gou2024critic} introduce an instructional framework where models revise explanations based on high-level critique from external tools. Similarly, \citet{perez2023discovering} use model-written evaluations to systematically probe LLM behaviors and identify failure modes, while \citet{tan2023selfcriticism} explore self-evaluation and alignment, where models assess their own outputs against criteria like helpfulness, honesty, and harmlessness. GIER differs from these methods by eliminating both instructive feedback and revision exemplars; models must independently infer both the flaws and the corrective actions with respect to given gap definitions.

\paragraph{Explanation Supervision and Evaluation.}
Explanation-based datasets probe whether model predictions are justified and interpretable. In natural language inference, e-SNLI \citep{camburu2018snli} and ERASER \citep{deyoung2020eraser} provide gold rationales to evaluate plausibility and sufficiency. In QA, datasets like CoS-E \citep{rajani2019explain}, ECQA \citep{aggarwal2021ecqa}, and StrategyQA \citep{geva2021strategyqa} incorporate multi-step or commonsense justifications, while HotpotQA \citep{yang2018hotpotqa} and QASC \citep{khot2020qasc} emphasize multi-hop reasoning. These resources are primarily used for post hoc evaluation or supervised rationale training. We instead use them to evaluate a new process-oriented paradigm: rather than assessing rationales post hoc, we guide models to iteratively refine them in response to specified gaps.

\paragraph{Alignment through Natural Language Supervision.}
A final strand of work focuses on aligning models with human preferences via reinforcement learning with human feedback (RLHF) \citep{ouyang2022training, bai2022training}, preference modeling \citep{rafailov2023dpo}, critiques \citep{bai2022constitutionalaiharmlessnessai}, or by using high-quality examples for supervised fine-tuning \citep{zhou2023lima}. Related approaches leverage language models to express subjective evaluations or uncertainty in natural language \citep{kuhn2023uncertainty, zheng2023llmasjudge}. GIER shares the goal of aligning model behavior with abstract human-centric standards but does so without reward tuning or label-based supervision. Instead, we test whether conceptual norms can shape model behavior through structured self-reflection and revision.

\vspace{0.2em}
In summary, GIER integrates elements from prompting, critique-based revision, and explanation-centered evaluation but introduces a novel gap-driven formulation. It operationalizes abstract reasoning flaws as interpretable prompts and uses these prompts to elicit iterative self-improvement without examples or rewards. This distinguishes GIER as a lightweight, generalizable framework for probing conceptual alignment and self-correction in language models. 

\begin{figure*}[t]
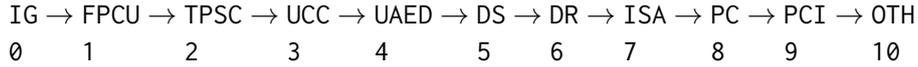

\centering
\verb+IG+ $\rightarrow$ \verb+FPCU+ $\rightarrow$ \verb+TPSC+ $\rightarrow$ \verb+UCC+ $\rightarrow$ \verb+UAED+ $\rightarrow$ \verb+DS+ $\rightarrow$ \verb+DR+ $\rightarrow$ \verb+ISA+ $\rightarrow$ \verb+PC+ $\rightarrow$ \verb+PCI+ $\rightarrow$ \verb+OTH+

\verb+0    1      2      3     4      5    6    7     8    9     10 +

\caption{A linear scaffold of privacy-related themes based on conceptual relevance. Numbers signify an integer representation of the concept type.}
\label{fig:unified_scaffold}
\end{figure*}

\section{Gap Definitions}
\label{apdx:gap_definitions}
The gap definitions we use in the three tasks are listed below.
\subsection{SciFact Rationale Generation}
\label{apdx:gap_definitions_scifact}

\textit{Coverage}: The rationale fails to accumulate all independent lines of reasoning from the source text that may lead to the decision in a stand-alone manner.

\noindent \textit{Conciseness}: The rationale includes supplemental explanations, details, or qualifiers that lead to a longer explanation.

\noindent \textit{Textual Grounding}: The rationale fails to anchor its reasoning in specific quotes or phrases from the source text.

\noindent \textit{Source Faithfulness}: The rationale introduces content, interpretations, or conclusions drawn from external knowledge that are not stated in the source text.

\noindent \textit{Unsupported Emphasis}: The rationale places undue weight on a piece of evidence, interpretation, or claim without sufficient justification or clear support from the source text.

\subsection{PrivacyQA Sentence Selection}
\label{apdx:gap_definitions_privacyqa}

\noindent \textit{Coverage}: The selection omits available information, relevant context, or important nuances, conditions, and exceptions required to fully answer the question.

\noindent \textit{Thematic Overreach}: The selection crosses into privacy-related themes that significantly widen the scope of the provided response.

\subsection{e-SNLI Commonsense Inference}
\label{apdx:gap_definitions_esnli}

\noindent \textit{Quantitative and Comparative Reasoning}: The output overlooks or misinterprets numeric or comparative information, including quantities, dates, durations, orderings (e.g., first, more, twice), or arithmetic relations.

\noindent \textit{Reference Resolution}: The output fails to correctly match pronouns, names, or noun phrases to the right entities across the context and statement.

\noindent \textit{Logical Inferences}: The output mishandles logical structures, such as negation, conditionals (if/then), conjunctions (and/or), or syntactic transformations (e.g., passive to active).

\noindent \textit{Pragmatic Inferences}: The output mishandles pragmatic inferences, including drawing incorrect conclusions from implicit facts, not accounting for possibilities left unclear in the context, or failing to apply commonsense reasoning based on the context. 

\noindent \textit{Lexical Inferences:} The output misses inferences based on word meaning (e.g., synonyms, antonyms, scalar terms) or lexical similarities.

\section{Privacy Thematic Drift}
\label{apdx:thematic_drift}

In evaluating question answering over privacy policies, a model tends to extract additional sentences beyond the gold standard set. While they degrade a model's precision, these extra sentences often  provide contextual or background information that supports comprehension. For example, when a question concerns data retention, a few sentences describing the collected data may appear first.

To quantitatively assess how far off the selected sentences deviate semantically from a question's focus, we use the \textsf{Thematic Drift} metric, which measures the average semantic distance of a set of sentences relative to the question type.

Let $U$ denote a \textit{unified scaffold}, a predefined proximity mapping of semantic types commonly found in privacy policies. For example, we use the linear scaffold illustrated in Figure~\ref{fig:unified_scaffold} for this work.
In this linear scaffold, each type is assigned an integer index, referred to as \textit{type-value}, based on its position in $U$. For a question of type-value $X$ and set of sentence type-values $S = \{Z_1, Z_2, \dots, Z_m\}$ we define:
\[
\textsf{Thematic Drift}(S) = \frac{1}{m} \sum_{j=1}^m |Z_j - X|
\]

The PrivacyQA dataset already has type labels for all questions. We use the XLNet based privacy policy sentence classifier proposed by \citet{adhikari2022sentence} to obtain the type labels for each individual sentence of a policy.

A thematic drift of zero signifies that all sentences in the selection is of the same type as the question. For a value $\approx 1$, the selection is semantically proximate to the question type. Consequently, the metric provides an alternative to lexical overlap and semantic similarity, and captures thematic focus and interpretive alignment of a selection of text to a privacy question. It is also possible to define the unified scaffold in a two or three-dimensional space for alternative representation of proximity.

\begin{table*}[t]
\begin{centering}
\begin{tabular}{crp{0.09\textwidth}R{0.08\textwidth}R{0.18\textwidth}p{0.15\textwidth}}
\hline 
\textbf{Task} & \textbf{Metric} & \textbf{Wilcoxon $p$-value} & \textbf{Mean $\Delta$} & \textbf{Bootstrap 95\% CI (mean $\Delta$)} & \textbf{Statistically Sig. $\Delta$?}\tabularnewline
\hline 
Scifact & Decision Accuracy & $1.00000$ & $+0.00$ & $[-0.03,+0.03]$ & No\tabularnewline
 & Rationales Recall & $0.00013$ & $+0.11$ & $[+0.07,+0.16]$ & Yes $\blacktriangle$  \tabularnewline
 & Grounding Ratio & $0.00005$ & $+0.05$ & $[+0.03,+0.07]$ & Yes $\blacktriangle$ \tabularnewline
\hline 
PrivacyQA & Decision Accuracy & $0.10829$ & $-0.07$ & $[-0.16,+0.01]$ & No\tabularnewline
 & Selection Precision & $0.00375$ & $-0.14$ & $[-0.23,-0.06]$ & Yes $\blacktriangledown$ \tabularnewline
 & Selection Recall & $0.00007$ & $+0.20$ & $[+0.13,+0.28]$ & Yes $\blacktriangle$ \tabularnewline
 & Thematic Drift & $0.00045$ & $+0.69$ & $[+0.33,+1.09]$ & Yes $\blacktriangle$ \tabularnewline
\hline 
e-SNLI & Decision Accuracy & $0.61708$ & $-0.02$ & $[-0.10,+0.06]$ & No\tabularnewline
 & Reasoning Attribution & $0.31731$ & $-0.04$ & $[-0.13,+0.04]$ & No\tabularnewline
\hline 
\end{tabular}
\par\end{centering}
\caption{Statistical significance analysis of performance improvements between
GIER initial and final iteration outputs. Wilcoxon $p$-values indicate
whether per-sample changes are significant. Mean $\Delta$ reflects the
average improvement (or decline) from initial to final iteration.
The bootstrap CI indicates the range in which this average difference
likely lies. Results are from the GPT-4.1 model.}

\label{tab:statistical_significance}
\end{table*}

\section{Prompts}
\label{apdx:prompts}
Figures~\ref{fig:scifact_initial_prompt}, \ref{fig:privacyqa_initial_prompt}, and \ref{fig:esnli_initial_prompt} show the GIER initial prompt used for the three tasks. Figures~\ref{fig:scifact_revision_prompt}, \ref{fig:privacyqa_revision_prompt}, and \ref{fig:esnli_revision_prompt} show the GIER revision prompt used for the three tasks. The highest input token size of 20,296 appears in iteration 5 for PrivacyQA, but is within the context window length of all models. Similarly, output token sizes are also within all model limits.

\begin{table*}[t]
\begin{centering}
\begin{tabular}{rp{0.15\textwidth}p{0.15\textwidth}l}
\hline 
\multicolumn{4}{c}{\textbf{SciFact rationale generation}}\tabularnewline
\hline 
 & \textit{Baseline \newline w/o gap defns.} & \textit{Baseline \newline w/ gap defns.} & \textit{GIER Iter. 1} $\rightarrow\cdots\rightarrow$ \textit{Iter. 5}\tabularnewline
Decision Accuracy & $0.91$ & $0.92$ & $0.92\rightarrow0.92\rightarrow0.92\rightarrow0.92\rightarrow0.92$\tabularnewline
Rationales Recall & $0.63$ & $0.74$ & $0.80\rightarrow0.84\rightarrow0.86\rightarrow0.86\rightarrow0.85$\tabularnewline
Grounding Ratio & $0.54$ & $0.58$ & $0.57\rightarrow0.59\rightarrow0.62\rightarrow0.63\rightarrow0.63$\tabularnewline
\hline 
\multicolumn{4}{c}{\textbf{PrivacyQA sentence selection}}\tabularnewline
\hline 
 & \textit{Baseline \newline w/o gap defns.} & \textit{Baseline \newline w/ gap defns.} & \textit{GIER Iter. 1} $\rightarrow\cdots\rightarrow$ \textit{Iter. 5}\tabularnewline
Decision Accuracy & $0.70$ & $0.70$ & $0.63\rightarrow0.63\rightarrow0.63\rightarrow0.63\rightarrow0.63$\tabularnewline
Selection Precision & $0.68$ & $0.68$ & $0.61\rightarrow0.57\rightarrow0.54\rightarrow0.54\rightarrow0.53$\tabularnewline
Selection Recall & $0.44$ & $0.45$ & $0.52\rightarrow0.58\rightarrow0.63\rightarrow0.65\rightarrow0.66$\tabularnewline
Thematic Drift & $0.67$ & $0.69$ & $1.07\rightarrow1.30\rightarrow1.39\rightarrow1.37\rightarrow1.36$\tabularnewline
\hline 
\multicolumn{4}{c}{\textbf{e-SNLI commonsense inference}}\tabularnewline
\hline 
 & \textit{Baseline \newline w/o gap defns.} & \textit{Baseline \newline w/ gap defns.} & \textit{GIER Iter. 1} $\rightarrow\cdots\rightarrow$ \textit{Iter. 5}\tabularnewline
Decision Accuracy & $0.57$ & $0.57$ & $0.56\rightarrow0.55\rightarrow0.55\rightarrow0.55\rightarrow0.55$\tabularnewline
Reasoning Attribution & $0.62$ & $0.75$ & $0.70\rightarrow0.69\rightarrow0.70\rightarrow0.73\rightarrow0.71$\tabularnewline
\hline 
\end{tabular}
\par\end{centering}
\caption{Quantitative metrics for three GIER tasks over iterations. Baseline with gap definitions is the GIER initial prompt. Results are from the GPT-4.1 model.}

\label{tab:iterations}
\end{table*}

\section{Statistical Significance}
\label{apdx:statistical_significance}

We evaluated the statistical significance of improvements observed from the initial GIER iteration (with gap definitions) to the final GIER iteration using the Wilcoxon signed-rank test on paired per-example metric values. Table~\ref{tab:statistical_significance} summarizes the results from the tests. For SciFact rationale recall, GIER yielded a significant gain from 0.74 to 0.85 ($p<0.001$, Wilcoxon). Bootstrap resampling with 10,000 iterations further confirmed this effect, producing a 95\% confidence interval for the mean recall improvement of $[+0.07,+0.16]$, indicating a robust and practically meaningful gain. Similar significant improvements were observed for grounding ratio ($p<0.001$, Wilcoxon; 95\% CI $[+0.03, +0.07]$) and PrivacyQA sentence selection recall ($p<0.001$, Wilcoxon; 95\% CI $[+0.13,+0.28]$). PrivacyQA selection precision showed a modest but statistically significant decrease from 0.68 to 0.53 ($p< 0.05$, Wilcoxon; 95\% CI $[-0.23,-0.06]$), reflecting a precision-recall trade-off consistent with thematic drift findings. Thematic drift increase is statistically significant ($p<0.001$, Wilcoxon), but CI upper limit of 1.09 on mean difference still suggests conceptual relevance of false positives. In e-SNLI, reasoning attribution stayed stable with $p>0.1$. Decision accuracy showed no significant change across all tasks ($p>0.1$).

These results demonstrate that iterative enhancement with GIER reliably improves reasoning-related metrics, with statistically significant and confidently bounded effect sizes. Since decision accuracy remains unchanged even when gap-definitions are introduced at the baseline stage, we infer that GIER acts less as a decision calibrator and more as an explanation refiner. This capability is particularly valuable in tasks where label accuracy saturates early, but rationale quality remains improvable.

\begin{sidewaystable*}
\begin{centering}
\begin{tabular}{rccccc}
\hline 
\multicolumn{6}{c}{\textbf{SciFact rationale generation}}\tabularnewline
\hline 
\\ 
 & \textit{GIER intial} $\rightarrow$ \textit{Iter. 1} & \textit{Iter. 1} $\rightarrow$ \textit{Iter. 2} & \textit{Iter. 2} $\rightarrow$ \textit{Iter. 3} & \textit{Iter. 3} $\rightarrow$ \textit{Iter. 4} & \textit{Iter. 4} $\rightarrow$ \textit{Iter. 5}\tabularnewline
Decision Accuracy & $0.00\ (p=1.000)$ & No change & No change & No change & No change\tabularnewline
Rationales Recall & \textbf{$\mathbf{+8.3\%\ (p=0.008)}$} & \textbf{$\mathbf{+5.4\%\ (p=0.030)}$} & $+1.7\%\ (p=0.343)$ & $+0.1\%\ (p=0.673)$ & $-0.9\%\ (p=0.588)$\tabularnewline
Grounding Ratio & $-2.4\%\ (p=0.372)$ & \textbf{$\mathbf{+4.7\%\ (p=0.002)}$} & \textbf{$\mathbf{+4.2\%\ (p<0.001)}$} & \textbf{$\mathbf{+2.0\%\ (p=0.042)}$} & $+0.1\%\ (p=0.404)$\tabularnewline
\\ 
\hline 
\multicolumn{6}{c}{\textbf{PrivacyQA sentence selection}}\tabularnewline
\hline 
\\ 
 & \textit{GIER intial} $\rightarrow$ \textit{Iter. 1} & \textit{Iter. 1} $\rightarrow$ \textit{Iter. 2} & \textit{Iter. 2} $\rightarrow$ \textit{Iter. 3} & \textit{Iter. 3} $\rightarrow$ \textit{Iter. 4} & \textit{Iter. 4} $\rightarrow$ \textit{Iter. 5}\tabularnewline
Decision Accuracy & $-9.9\%\ (p=0.108)$ & No change & No change & No change & No change\tabularnewline
Selection Precision & $\mathbf{-13.3\%\ (p=0.026)}$ & $\mathbf{-7.4\%\ (p=0.029})$ & $\mathbf{-3.9\%\ (p=0.023)}$ & $-1.5\%\ (p=0.076)$ & $-0.4\%\ (p=0.445)$\tabularnewline
Selection Recall & $\mathbf{+16.6\%\ (p=0.009)}$ & $\mathbf{+13.1\%\ (p=0.001)}$ & $\mathbf{+7.5\%\ (p<0.001)}$ & $\mathbf{+2.9\%\ (p=0.012)}$ & $\mathbf{+1.8\%\ (p=0.035)}$\tabularnewline
Thematic Drift & $\mathbf{+62.7\%\ (p=0.012)}$ & $+21.4\%\ (p=0.076)$ & $+7.1\%\ (p=0.061)$ & $-1.4\%\ (p=0.124)$ & $\mathbf{-1.1\%\ (p=0.043)}$\tabularnewline
\\ 
\hline 
\multicolumn{6}{c}{\textbf{e-SNLI commonsense inference}}\tabularnewline
\hline 
\\ 
 & \textit{GIER intial} $\rightarrow$ \textit{Iter. 1} & \textit{Iter. 1} $\rightarrow$ \textit{Iter. 2} & \textit{Iter. 2} $\rightarrow$ \textit{Iter. 3} & \textit{Iter. 3} $\rightarrow$ \textit{Iter. 4} & \textit{Iter. 4} $\rightarrow$ \textit{Iter. 5}\tabularnewline
Decision Accuracy & $-1.8\%\ (p=0.796)$ & $-1.8\%\ (p=0.317)$ & No change & No change & No change\tabularnewline
Reasoning Attribution & $-5.7\%\ (p=0.343)$ & $-2.0\%\ (p=0.564)$ & $+1.0\%\ (p=0.655)$ & $+4.0\%\ (p=0.102)$ & $-2.9\%\ (p=0.180)$\tabularnewline
\\ 
\hline 
\end{tabular}
\par\end{centering}
\caption{Mean relative change and Wilcoxon paired test $p$-values for three GIER tasks over iterations. Bold values imply a statistically significant change ($p<0.05$). `No change' implies the values are identical across the samples in the two iterations. Results are from the GPT-4.1 model. Note: Mean relative change is computed from full-precision values while Table~\ref{tab:iterations} reports rounded numbers. Wilcoxon paired test in PrivacyQA done on pairs where both had an answerable label.}

\label{tab:iterations_sig_test}
\end{sidewaystable*}

\begin{sidewaystable*}
\begin{centering}
\begin{tabular}{p{0.22\textwidth}p{0.07\textwidth}p{0.07\textwidth}p{0.07\textwidth}p{0.07\textwidth}l}
\hline 
\multicolumn{6}{c}{\textbf{SciFact rationale generation}}\tabularnewline
\hline 
\textit{Ablation Type} & \textit{Decision Accuracy} & \textit{Rationales Recall} & \textit{Grounding Ratio} &  & \textit{Comment}\tabularnewline
Coverage Drop & $-1.0\%$ & $-23.5\%$ & $+9.5\%$ &  & Key driver of multi-evidence recall\tabularnewline
Conciseness Drop & $-2.1\%$ & $+4.7\%$ & $-11.1\%$ &  & Controls verbosity and grounding\tabularnewline
Reflection Drop & No Change & $-8.2\%$ & $-19.0\%$ &  & Enhances both grounding and completeness\tabularnewline
\hline 
\multicolumn{6}{c}{\textbf{PrivacyQA sentence selection}}\tabularnewline
\hline 
\textit{Ablation Type} & \textit{Decision Accuracy} & \textit{Selection Precision }& \textit{Selection Recall} & \textit{Thematic Drift} & \textit{Comment}\tabularnewline
Coverage Drop & $+14.3\%$ & $+35.8\%$ & $-48.5\%$ & $-63.2\%$ & Key driver of broader interpretation and retrieval\tabularnewline
Thematic Overreach Drop & No change & $-5.7\%$ & $+3.0\%$ & $+7.4\%$ & Soft-constraint to suppress conceptual drift\tabularnewline
Reflection Drop & $+4.8\%$ & $+32.1\%$ & $-16.7\%$ & $-51.5\%$ & Secondary driver of broader interpretation and retrieval\tabularnewline
\hline 
\multicolumn{6}{c}{\textbf{e-SNLI commonsense inference}}\tabularnewline
\hline 
\textit{Ablation Type} & \textit{Decision Accuracy} & \textit{Reasoning Attribution} &  &  & \textit{Comment}\tabularnewline
Pragmatic Inferences Drop & $+1.8\%$ & $-1.4\%$ &  &  & Low impact gap in the presence of others\tabularnewline
Logical \& Lexical Inferences Drop & $+1.8\%$ & $-7.0\%$ &  &  & Moderate contribution to reasoning discovery\tabularnewline
Reflection Drop & $+9.0\%$ & $-22.5\%$ &  &  & Key driver to retain identified reasoning chunks across iterations\tabularnewline
\hline 
\end{tabular}
\par\end{centering}
\caption{Relative changes in key metrics at the final GIER iteration when specific gap definitions or gap-based reflection steps are omitted during revision. Results are from the GPT-4.1 model.}

\label{tab:ablation_drop}
\end{sidewaystable*}

\section{Iteration Variations}
\label{apdx:iteration_variations}
We evaluated GIER's performance over five post-initialization iterations to assess whether additional refinement cycles yield sustained improvements. Table~\ref{tab:iterations} reports metric changes at each iteration and Table~\ref{tab:iterations_sig_test} shows statistical significance results based on Wilcoxon signed-rank paired tests.

Across tasks, iterations 1–3 account for the majority of gains, with subsequent changes becoming negligible or reversing. In SciFact, rationale recall improves significantly from GIER initial to iteration 3 ($+8.3\%, +5.4\%, +1.7\%$), but then flattens ($+0.1\%$) and declines ($-0.9\%$). Grounding ratio increases through iteration 4 but stalls afterward. Decision accuracy remains unchanged throughout, indicating stable label predictions despite rationale edits. In PrivacyQA, sentence recall improves at every step, but with diminishing returns. Notably, sentence precision degrades steadily over the same period, suggesting a tradeoff between recall and focus. Thematic drift has the highest jump of $+62.7\%$ in the initial prompt, while later stages show statistically insignificant changes. This indicates that the most significant changes often happen in the first round of revision. In e-SNLI, the largest improvement occurs during the shift from a generic prompt to GIER initial (Table~\ref{tab:iterations}). All iterations beyond that produce fluctuations in attribution and accuracy metrics without statistical significance. This suggests that the conceptual guidance embedded in the GIER prompt is sufficient for improvement in this task, with iterations offering no consistent value.

These trends support our practical decision to limit GIER to five iterations post-initialization. By iteration 3, most metrics plateau, with gains becoming smaller and $p$-values increasing. No metric exhibits late-stage inflection or recovery, indicating absence of latent improvements. From a behavioral perspective, this mirrors human editing cycles where major issues are corrected early, while later revisions tend to focus on marginal tweaks or risk overcorrection. 

\section{Ablation Study}
\label{apdx:ablation}
To assess the impact of gap definitions and explicit self-analysis on performance, we perform two ablation variants: (1) gap drop, and (2) reflection drop. Gap drop follows the same GIER procedure but with a reduced set of gap definitions. Reflection drop retains the full set of gaps but solicits revisions without gap-specific scoring or explanation. Table~\ref{tab:ablation_drop} summarizes the relative changes in the metric values for different ablation variants. Additionally, Appendix~\ref{apdx:iteration_variations} explores iteration-based variants, which serve a similar role to an iteration drop ablation.

\paragraph{SciFact.}
Gap drop ablation in SciFact highlights distinct roles for the coverage and conciseness gaps in shaping rationale quality. Relative to the full setup with both gaps active, dropping the coverage constraint produces a substantial reduction in rationale recall ($-23.5\%$), indicating its central role in identifying multiple evidences. This comes with a moderate increase in grounding ratio ($+9.5\%$), as removing the pressure for coverage also leads to reduced synthesized content to integrate the multiple rationales. The impact on decision accuracy is minor ($-1.0\%$), implying that coverage primarily affects the breadth of evidence without substantially altering the final label decision. In contrast, dropping the conciseness gap leads to a drop in grounding ratio ($-11.1\%$) and a small gain in rationale recall ($+4.7\%$), indicating that conciseness functions as a key constraint on verbosity and plays an important role in maintaining textual grounding. Decision accuracy again shows only a small reduction ($-2.1\%$), reinforcing that these constraints predominantly shape the form and quality of rationales rather than the final label. Dropping reflection leads to reductions in both rationale recall and grounding ratio, confirming that it plays a critical role in preserving completeness and source alignment.

\paragraph{PrivacyQA.}
Gap drop ablation in PrivacyQA also reveals distinct roles for the two gap definitions: coverage and thematic overreach. Compared to the case when both gaps are active, dropping the coverage gap yields a substantial increase in decision accuracy ($+14.3\%$), i.e., the model becomes better at correctly determining whether a question is answerable or not. This comes alongside a large gain in selection precision ($+35.8\%$) but a sharp reduction in selection recall ($-48.5\%$) and thematic drift ($-63.2\%$). This suggests that enforcing coverage pushes the model to broader interpretation of questions, as well as retrieving broader content. Without the coverage constraint, the model adopts a conservative, high-precision strategy that favors abstaining when uncertain, even at the cost of omitting some relevant details. By contrast, dropping the thematic overreach constraint has only modest effects: a slight reduction in precision ($-5.7\%$), a small gain in recall ($+3.0\%$), and a mild increase in thematic drift ($+7.4\%$), with no change in decision accuracy. This indicates that thematic overreach serves mainly as a soft constraint to reduce peripheral content but has little influence on the model's core ability to determine answerability. Reflection drop results follow a similar trend as in coverage drop, albeit with a milder effect, suggesting that gap-specific explanations play a vital role in broader interpretation of questions. 

\paragraph{e-SNLI.}
Ablation experiments in e-SNLI reveal that both gap-specific scaffolding and reflection play critical roles in shaping reasoning alignment. Dropping the pragmatic inference gap produces a modest relative gain in decision accuracy ($+1.8\%$) and a slight reduction in reasoning attribution ($-1.4\%$); the reasoning chunks covered under this gap are potentially also identified in other gaps. In contrast, dropping the logical and lexical inference gaps yield the same gain in accuracy ($+1.8\%$) but a larger decline in reasoning attribution ($-7.0\%$), indicating that this gap plays a more substantial role in identifying human reasoning chunks. The most pronounced effects emerge from the reflection drop ablation: removing reflection leads to a substantial improvement in decision accuracy ($+9.0\%$) but a sharp decline in reasoning attribution ($-22.5\%$). Recall that the GIER initial prompt recovers a substantial portion of the human reasoning chunks (0.75, Table~\ref{tab:result_summary}); the significant drop in this value without reflection demonstrates that explicit gap-based analyses during revisions are critical to avoid the loss of reasoning pathways. 

\paragraph{Overall.}
The ablation results reveal consistent patterns in how gap scaffolding shapes model behavior. Across tasks, gap-based analyses and self-reflection consistently serve as a key driver of explanation completeness. The influence of individual gaps is more task-specific, functioning either as strong behavioral drivers---shaping retrieval breadth, grounding, or reasoning discovery---or as soft regularizers that constrain peripheral content. More broadly, the results demonstrate that conceptual scaffolding in GIER is not merely a tool for post hoc explanation improvement but an active mechanism that steers the model's reasoning strategy, retrieval behavior, and even its decision-making stance according to task demands.

\section{Reflections on Gap Writing}
\label{apdx:reflections_gap_writing}

During our pilot study with SciFact, 
we initially hand-crafted a set of eight gap criteria to diagnose and improve LLM-generated rationales. 
Each gap was designed to capture a distinct failure mode in rationales, with the aim of promoting nuanced, multi-dimensional self-revision. However, through iterative experiments, we observed that while these categories were conceptually rigorous, they did not consistently elicit the intended behaviors from the model. 
We found that a smaller set of more behaviorally grounded and directive gaps produced significantly better outcomes. We draw several methodological lessons from this transition:

\begin{itemize}
    \item Gaps that align with easily observable behaviors are more reliably acted upon by LLMs.

    \item Naming of gaps influence behavior, and definitions should target a specific failure pattern that the LLM can spot. 

    \item Fewer, sharper criteria work better than broad taxonomies. 

    \item Longer gap definitions or overlapping gaps made the prompt harder to effectively act on.
\end{itemize}

This refinement reflects a broader theme in prompt engineering and LLM alignment---the most effective abstractions are those that LLMs can act on, not merely recognize.

\section{Extended Discussion}
\label{apdx:extended_discussion}

\paragraph{PrivacyQA Answering Unanswerable Questions.}

Out of 50 unanswerable questions in PrivacyQA, GPT-4.1 produced answers for 33 of them from the first GIER iteration onward, compared to 21 with the initial prompt. A manual analysis of the 33 samples reveals that 11 were clearly poor or irrelevant, while 9 included supporting sentences that were at least topically related to the question. The remaining cases reflect broad or reinterpretive reasoning. For instance, the question \textit{``Who is the accounting information going to?''} is treated as a third-party sharing query, prompting a plausible answer set about data disclosures to external entities. One of the questions is logically impossible to answer---\textit{``Is anything not mentioned installed on my phone?''}---but the model reframes it as a first-party data collection query. These patterns suggest that GIER's revision process can promote flexible reinterpretation when questions are topically extensible, but does not universally override unanswerability.

\paragraph{e-SNLI Gap Role in Reasoning Attribution.}
\begin{figure}[t]
    \centering
    \includegraphics[scale=0.40]{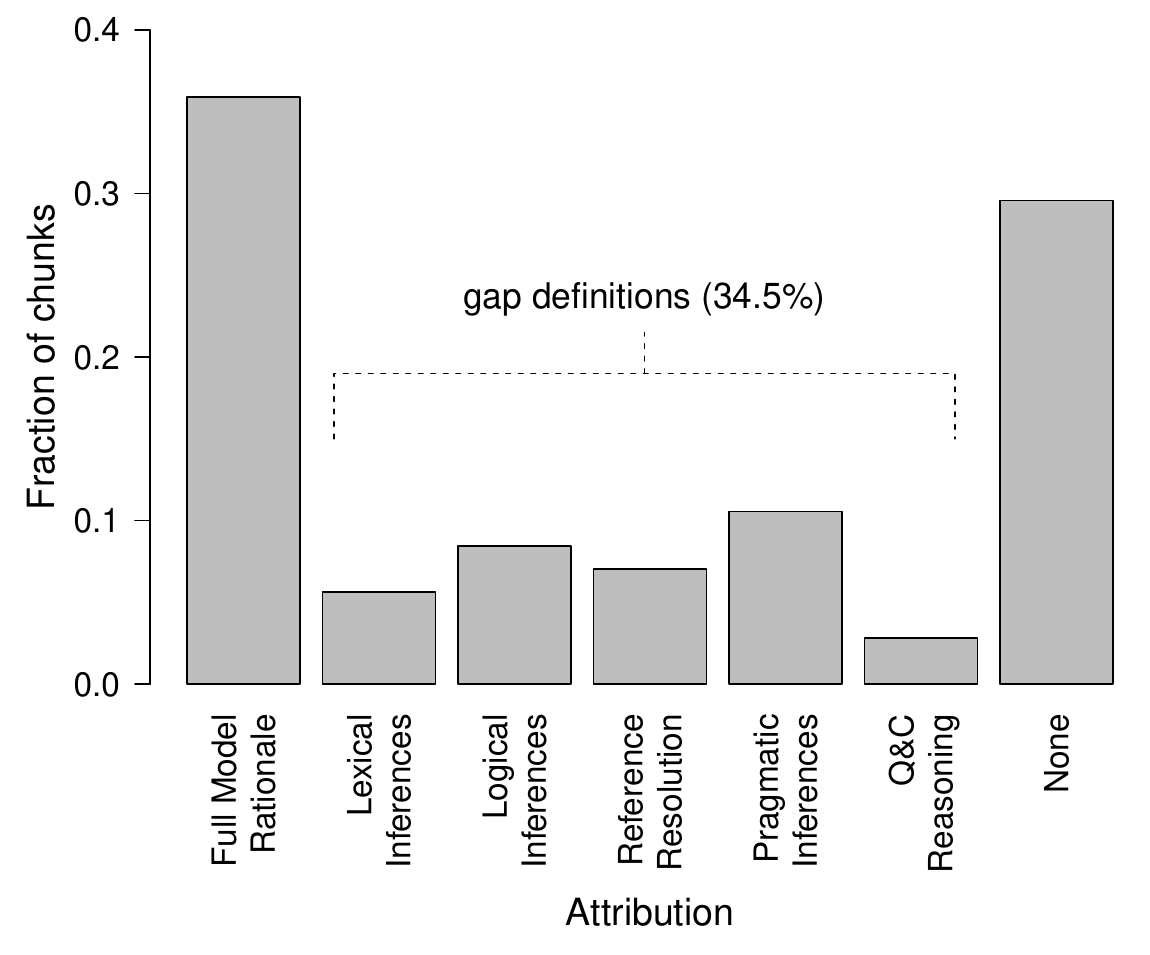}
    \caption{Fraction of gold reasoning chunks across e-SNLI samples that are best attributed (highest entailment or contradiction score > 0.8) to the full model reasoning, a gap explanation, or none in GIER final iteration. There are 142 total chunks. Gap explanations make up for 34.5\% of attribution. Results are from the GPT-4.1 model}
    \label{fig:esnli_gap_attribution}
\end{figure}

Figure~\ref{fig:esnli_gap_attribution} shows how gold reasoning chunks in e-SNLI were attributed in the final GIER iteration (GPT-4.1), either to the full-model rationale, to a specific gap explanation, or to neither. Approximately one-third of the gold reasoning chunks were best attributed to model explanations produced during gap analysis. The pragmatic inferences gap accounted for about $11\%$ of the chunks, followed by logical inferences at $9\%$, consistent with the inferential demands of e-SNLI. Although GIER's initial prompt included gap definitions, gap-based self-analysis occurred only during the revision steps. While the overall reasoning attribution score (Table~\ref{tab:iterations}) remained relatively stable across iterations, this finer-grained view reveals which conceptual gaps most directly facilitated alignment with human reasoning trajectories.

\paragraph{Model Intent vs. Actions.}
To assess whether a model's revised outputs reflect the revision steps it intended based on gap analysis, we manually reviewed GPT-4.1's initial GIER outputs, the consolidated gap explanations, and the corresponding revisions. Each instance was labeled as \verb+none+ (no improvements suggested and none performed), \verb+aligned+ (all suggested improvements were performed), \verb+partial+ (some performed), or \verb+misaligned+ (none performed despite suggestions). We observed strong alignment between the gap analysis and the revisions, with no \verb+misaligned+ cases. In SciFact, all instances involved suggested improvements, all fully \verb+aligned+. PrivacyQA had 11 \verb+none+, 81 \verb+aligned+, and 8 \verb+partial+ cases, typically when no appropriate sentences were available to fully implement the suggested revisions. e-SNLI had the most \verb+none+ labels (27), indicating no revisions were needed, while 70 were fully \verb+aligned+. These results show that the model's revisions consistently follow its own gap analyses, demonstrating high intent-action consistency.

\paragraph{Decision Revisions.}
A common concern with rationale-focused methods is that models may rationalize their initial prediction without reconsidering the decision itself. To test whether GIER prompts genuine decision revision, we tracked label changes from the initial to final iterations when using GPT-4.1. Across all three tasks, the model did revise decisions, indicating that GIER is not purely a post hoc justification mechanism. In e-SNLI, 19 decisions changed---7 correct, 12 incorrect---demonstrating a capacity for revision. As discussed earlier, even incorrect decisions often reflected reasoning that aligned with or thoughtfully challenged the gold rationale. SciFact, with high initial accuracy, saw only two switches (one correct, one incorrect). PrivacyQA also had 19 switches, 13 of them incorrect, with 12 of those reflecting shifts from ``unanswerable'' to ``answerable,'' often due to broad reinterpretations.

\section{Qualitative Examples}
\label{apdx:qualitative_examples}

The following examples are based on the GIER iterations of GPT-4.1. Referred baseline model is the one without any gap definitions.

\subsection{SciFact}

\begin{example}{Multi-evidence rationale surfaced by GIER (italicized text = quoted segment; highlighted text = in some gold evidence set).}
\textbf{Abstract}: Uncoupling protein 1 (UCP1) plays a central role in nonshivering thermogenesis in brown fat; however, its role in beige fat remains unclear. Here we ... Our study uncovers a noncanonical thermogenic mechanism through which beige fat controls whole-body energy homeostasis via Ca2+ cycling.

\textbf{Claim}: Ca2+ cycling is a UCP1-dependent thermogenic mechanism.

\textbf{Gold Decision}: \textit{refute}

\textbf{Gold Evidence A}: Here we report a robust UCP1-independent thermogenic mechanism in beige fat that involves enhanced ATP-dependent Ca2+ cycling by sarco/endoplasmic reticulum Ca2+-ATPase 2b (SERCA2b) and ryanodine receptor 2 (RyR2).

\textbf{Gold Evidence B}: Conversely, enhanced Ca2+ cycling by activation of $\alpha$1- and/or $\beta$3-adrenergic receptors or the SERCA2b-RyR2 pathway stimulates UCP1-independent thermogenesis in beige adipocytes.
\\

\textbf{Model Baseline Decision:} \textit{refute}$\checkmark$

\textbf{Model Baseline Reason}: The text states that there is a \hl{\textit{`robust UCP1-independent thermogenic mechanism in beige fat that involves enhanced ATP-dependent Ca2+ cycling by sarco/endoplasmic reticulum Ca2+-ATPase 2b (SERCA2b) and ryanodine receptor 2 (RyR2).'}}$_\mathsf{Evidence\  A}$ This indicates that Ca2+ cycling is not dependent on UCP1 for thermogenesis.
\\

\textbf{Model Final Label}: \textit{refute}$\checkmark$

\textbf{Iter 1 Coverage Explanation}: The rationale covers the main lines of evidence: that Ca2+ cycling is UCP1-independent and that this mechanism is present in species lacking UCP1. However, it does not explicitly mention the activation of Ca2+ cycling via adrenergic receptors or the metabolic consequences (e.g., glucose sink), both of which further support the independence from UCP1.

\textbf{Model Final Reason}: The claim is refuted because the text explicitly states that Ca2+ cycling is a UCP1-independent thermogenic mechanism. The text describes \hl{\textit{`a robust UCP1-independent thermogenic mechanism in beige fat that involves enhanced ATP-dependent Ca2+ cycling by sarco/endoplasmic reticulum Ca2+-ATPase 2b (SERCA2b) and ryanodine receptor 2 (RyR2).'}}$_\mathsf{Evidence\ A}$ It provides evidence from species lacking UCP1: \textit{`inhibition of SERCA2b impairs UCP1-independent beige fat thermogenesis in humans and mice as well as in pigs, a species that lacks a functional UCP1 protein.'} The activation of this mechanism is supported by the statement: \hl{\textit{`enhanced Ca2+ cycling by activation of $\alpha$1- and/or $\beta$3-adrenergic receptors or the SERCA2b-RyR2 pathway stimulates UCP1-independent thermogenesis in beige adipocytes.'}}$_\mathsf{Evidence\  B}$ The metabolic consequences are described as: \textit{`beige fat thereby functions as a glucose sink and improves glucose tolerance independently of body weight loss.'} These statements collectively demonstrate that Ca2+ cycling is not UCP1-dependent, directly refuting the claim.

\end{example}

\begin{example}{Gold evidence set with two sentences surfaced by GIER (italicized text = quoted segment; highlighted text = in some gold evidence set).}
\textbf{Abstract}: Maintaining hematopoietic stem cell (HSC) quiescence is a critical property for the life-long generation of blood cells. Approximately 75\% of cells ... Therefore, GATA-3 is required for the maintenance of a normal number of LT-HSCs and for their entry into the cell cycle.

\textbf{Claim}: GATA3 regulates cell cycle progression in bone marrow hematopoietic stem cells.

\textbf{Gold Decision}: \textit{support}

\textbf{Gold Evidence A}: Furthermore, Gata3 mutant hematopoietic progenitor cells fail to be recruited into an increased cycling state after 5-fluorouracil-induced myelosuppression. Therefore, GATA-3 is required for the maintenance of a normal number of LT-HSCs and for their entry into the cell cycle.
\\

\textbf{Model Baseline Decision:} \textit{support}$\checkmark$

\textbf{Model Baseline Reason}: The text states that \textit{`fewer Gata3-null LT-HSCs are in cycle'} and that \hl{\textit{`Gata3 mutant hematopoietic progenitor cells fail to be recruited into an increased cycling state after 5-fluorouracil-induced myelosuppression.'}}$_\mathsf{Evidence\  A}$ These findings indicate that GATA3 is necessary for normal cell cycle entry and progression in bone marrow hematopoietic stem cells, supporting the claim that GATA3 regulates cell cycle progression in these cells.
\\

\textbf{Model Final Label}: \textit{support}$\checkmark$

\textbf{Iter 1 Coverage Explanation}: The rationale covers the main lines of evidence: the reduction in LT-HSCs, fewer cells in cycle, failure to increase cycling after myelosuppression, and the concluding statement about GATA-3's requirement. However, it could more explicitly mention that the text specifies the context as bone marrow HSCs and clarify that the evidence is specific to LT-HSCs, not all HSCs or progenitors.

\textbf{Iter 2 Textual Grounding Explanation}: The rationale uses direct quotes from the text and attributes them to the findings, but could improve by quoting from single sentences only and by more clearly indicating the source of each quote. There is some minor paraphrasing that could be replaced with direct quotes.

\textbf{Model Final Reason}: The claim is supported. The text states, in one sentence, that \textit{`Gata3-null mutant mice generate fewer LT-HSCs and that fewer Gata3-null LT-HSCs are in cycle.'} Another sentence reports, \hl{\textit{`Gata3 mutant hematopoietic progenitor cells fail to be recruited into an increased cycling state after 5-fluorouracil-induced myelosuppression.'}}$_\mathsf{Evidence\  A}$ The text also concludes, in a separate sentence, \hl{\textit{`GATA-3 is required for the maintenance of a normal number of LT-HSCs and for their entry into the cell cycle.'}}$_\mathsf{Evidence\  A}$ These findings refer specifically to bone marrow long-term repopulating HSCs (LT-HSCs), and the text notes in another sentence that \textit{`a role for GATA-3 in any prethymic progenitor cell has not been established.'} Therefore, the evidence demonstrates that GATA3 regulates cell cycle progression in bone marrow LT-HSCs.

\end{example}

\begin{example}{Incorrect baseline conclusion corrected during GIER revision (italicized text = quoted segment; highlighted text = in some gold evidence set).}
\textbf{Abstract}: BACKGROUND The associations of blood pressure with the different manifestations of incident cardiovascular disease in a contemporary population have not been compared. In this study, ... These findings emphasise the need for new blood pressure-lowering strategies, and will help to inform the design of randomised trials to assess them. FUNDING Medical Research Council, National Institute for Health Research, and Wellcome Trust.

\textbf{Claim}: Decreased diastolic blood pressure (DBP) is associated with abdominal aortic aneurysm.

\textbf{Gold Decision}: \textit{refute}

\textbf{Gold Evidence A}: Compared with diastolic blood pressure, raised systolic blood pressure had a greater effect on angina, myocardial infarction, and peripheral arterial disease, whereas raised diastolic blood pressure had a greater effect on abdominal aortic aneurysm than did raised systolic pressure.
\\

\textbf{Model Baseline Decision:} \textit{support}$\times$

\textbf{Model Baseline Reason}: The text states that `raised diastolic blood pressure had a greater effect on abdominal aortic aneurysm than did raised systolic pressure,' and also that \textit{`pulse pressure associations were inverse for abdominal aortic aneurysm (HR per 10 mm Hg 0.91 [95\% CI 0.86-0.98])'}. Since pulse pressure is the difference between systolic and diastolic pressure, an inverse association suggests that lower diastolic blood pressure (for a given systolic) is associated with higher risk of abdominal aortic aneurysm. Therefore, decreased diastolic blood pressure is associated with abdominal aortic aneurysm.
\\

\textbf{Model Final Label}: \textit{refute}$\checkmark$

\textbf{Iter 1 Source Faithfulness Explanation}: The rationale introduces an interpretation that as pulse pressure increases (due to decreased DBP), risk decreases, but the text does not explicitly state that decreased DBP is associated with abdominal aortic aneurysm. The text actually suggests that higher DBP is associated with higher risk, so the rationale's conclusion is not fully faithful to the source.

\textbf{Iter 1 Unsupported Emphasis Explanation}: The rationale places significant emphasis on the inverse association of pulse pressure with abdominal aortic aneurysm to support the claim, but this is not clearly linked to decreased DBP in the text. The rationale does not sufficiently justify why this evidence should be weighted more heavily than the direct statement about raised DBP.

\textbf{Model Final Reason}: The claim that `decreased diastolic blood pressure (DBP) is associated with abdominal aortic aneurysm' is refuted by the text. The text states that \hl{\textit{`raised diastolic blood pressure had a greater effect on abdominal aortic aneurysm than did raised systolic pressure',}}$_\mathsf{Evidence\  A}$ indicating that higher, not lower, DBP is associated with increased risk. The text also reports that \textit{`pulse pressure associations were inverse for abdominal aortic aneurysm (HR per 10 mm Hg 0.91 [95\% CI 0.86-0.98])'}. However, this finding does not establish that decreased DBP is associated with increased risk, as the text does not provide evidence for such a relationship. Therefore, the evidence in the text does not support the claim.

\end{example}

\begin{example}{Decision contradicting the gold label (italicized text = quoted segment; highlighted text = in some gold evidence set).}
\textbf{Abstract}: Inflammasomes are multiprotein complexes that include members of the NLR (nucleotide-binding domain leucine-rich repeat containing) family and caspase-1. Once bacterial molecules ... However, caspase-11 was dispensable for the fusion of lysosomes with phagosomes containing nonpathogenic bacteria, uncovering a fundamental difference in the trafficking of phagosomes according to their cargo.

\textbf{Claim}: Active caspase-11 protein promotes pyroptosis.

\textbf{Gold Decision}: \textit{support}

\textbf{Gold Evidence A}: Here, we demonstrated that caspase-11 was dispensable for caspase-1 activation in response to Legionella, Salmonella, Francisella, and Listeria.

\textbf{Gold Evidence B}: We also determined that active mouse caspase-11 was required for restriction of L. pneumophila infection.
\\

\textbf{Model Baseline Decision:} \textit{refute}$\times$

\textbf{Model Baseline Reason}: The provided text does not mention pyroptosis or describe caspase-11 as promoting pyroptosis. Instead, it states that \hl{\textit{`active mouse caspase-11 was required for restriction of L. pneumophila infection'}}$_\mathsf{Evidence\  B}$ and that \textit{`caspase-11 promoted the fusion of the L. pneumophila vacuole with lysosomes by modulating actin polymerization through cofilin.'} There is no evidence in the text supporting the claim that active caspase-11 protein promotes pyroptosis.
\\

\textbf{Model Final Label}: \textit{refute}$\times$

\textbf{Iter 5 Consolidated Explanation}: Across all iterations, the rationale consistently refutes the claim by noting the absence of pyroptosis and summarizing the described functions of caspase-11 and its human homologs. To further improve, the revised rationale should explicitly state that all described roles of caspase-11 and its homologs are unrelated to pyroptosis, quote the sentence about caspase-11 being dispensable for lysosome fusion with nonpathogenic bacteria as a separate quote, and ensure all quotes are from single sentences. Conciseness is strong, but minor streamlining is possible. Textual grounding can be improved by quoting all relevant sentences. Source faithfulness and balance are well maintained and should be preserved.

\textbf{Model Final Reason}: The claim that active caspase-11 protein promotes pyroptosis is not supported by the text. The text does not mention ``pyroptosis'' or describe any form of cell death induced by caspase-11. Instead, it states that \hl{\textit{`active mouse caspase-11 was required for restriction of L. pneumophila infection.'}}$_\mathsf{Evidence\  B}$ It also states that \textit{`caspase-11 promoted the fusion of the L. pneumophila vacuole with lysosomes by modulating actin polymerization through cofilin.'} Importantly, the text notes that \hl{\textit{`caspase-11 was dispensable for caspase-1 activation in response to Legionella, Salmonella, Francisella, and Listeria.'}}$_\mathsf{Evidence\  A}$ Since caspase-1 activation is associated with pyroptosis, this suggests caspase-11 is not required for pyroptosis in these contexts. The text further states, \textit{`caspase-11 was dispensable for the fusion of lysosomes with phagosomes containing nonpathogenic bacteria, uncovering a fundamental difference in the trafficking of phagosomes according to their cargo.'} Additionally, it says that \textit{`human caspase-4 and caspase-5, homologs of mouse caspase-11, cooperated to restrict L. pneumophila infection in human macrophages.'} None of these described functions are related to pyroptosis, and the text provides no evidence that caspase-11 promotes this process.

\end{example}

\begin{example}{Emergence of connective inferences from GIER initial to final iteration (italicized text = quoted segment; highlighted text = in some gold evidence set; bold text = summative judgement).}
\textbf{Abstract}: CONTEXT Dexamethasone is widely used to prevent postoperative nausea and vomiting (PONV) in pediatric tonsillectomy. ... CONCLUSION In this study of children undergoing tonsillectomy, dexamethasone decreased the risk of PONV dose dependently but was associated with an increased risk of postoperative bleeding. TRIAL REGISTRATION clinicaltrials.gov Identifier: NCT00403806.

\textbf{Claim}: Dexamethasone decreases risk of postoperative bleeding.

\textbf{Gold Decision}: \textit{refute}

\textbf{Gold Evidence A}: Two of 53 (4\%; 95\% CI, 0.5\%-13\%) children who received placebo had bleeding compared with 6 of 53 (11\%; 95\% CI, 4\%-23\%), 2 of 51 (4\%; 95\% CI, 0.5\%-13\%), and 12 of 50 (24\%; 95\% CI, 13\%-38\%) who received dexamethasone at 0.05, 0.15, and 0.5 mg/kg, respectively (P = .003).

\textbf{Gold Evidence B}: Dexamethasone, 0.5 mg/kg, was associated with the highest bleeding risk (adjusted relative risk, 6.80; 95\% CI, 1.77-16.5).

\textbf{Gold Evidence C}: CONCLUSION In this study of children undergoing tonsillectomy, dexamethasone decreased the risk of PONV dose dependently but was associated with an increased risk of postoperative bleeding.
\\

\textbf{Model Baseline\footnote{This baseline is with gap definitions.} Decision:} \textit{refute}$\checkmark$

\textbf{Model Baseline\footnotemark[\value{footnote}] Reason}: The text states that \hl{\textit{`dexamethasone, 0.5 mg/kg, was associated with the highest bleeding risk (adjusted relative risk, 6.80; 95\% CI, 1.77-16.5).'}}$_\mathsf{Evidence\  B}$ It also reports that \textit{`there were 26 postoperative bleeding episodes in 22 children,'} and that \textit{`eight children had to undergo emergency reoperation because of bleeding, all of whom had received dexamethasone.'} Furthermore, the conclusion says, \hl{\textit{`dexamethasone decreased the risk of PONV dose dependently but was associated with an increased risk of postoperative bleeding.'}}$_\mathsf{Evidence\  C}$ These statements directly contradict the claim that dexamethasone decreases risk of postoperative bleeding.
\\

\textbf{Model Final Label}: \textit{refute}$\checkmark$

\textbf{Model Final Reason}: The claim is refuted by the text. The study reports, \hl{\textit{`dexamethasone, 0.5 mg/kg, was associated with the highest bleeding risk (adjusted relative risk, 6.80; 95\% CI, 1.77-16.5).'}}$_\mathsf{Evidence\  B}$ The text states, \textit{`There were 26 postoperative bleeding episodes in 22 children.'} For bleeding rates, the text reports: \hl{\textit{`Two of 53 (4\%; 95\% CI, 0.5\%-13\%) children who received placebo had bleeding.'}}$_\mathsf{Evidence\  A}$ It also states, \hl{\textit{`6 of 53 (11\%; 95\% CI, 4\%-23\%)'}}$_\mathsf{Evidence\  A}$ for dexamethasone at 0.05 mg/kg, \hl{\textit{`2 of 51 (4\%; 95\% CI, 0.5\%-13\%)'}}$_\mathsf{Evidence\  A}$ at 0.15 mg/kg, and \hl{\textit{`12 of 50 (24\%; 95\% CI, 13\%-38\%)'}}$_\mathsf{Evidence\  A}$ at 0.5 mg/kg. The text notes, \textit{`Eight children had to undergo emergency reoperation because of bleeding, all of whom had received dexamethasone.'} The text also reports, \hl{\textit{`P = .003'}}$_\mathsf{Evidence\  A}$ for the difference in bleeding rates. The conclusion states, \hl{\textit{`dexamethasone decreased the risk of PONV dose dependently but was associated with an increased risk of postoperative bleeding.'}}$_\mathsf{Evidence\  C}$ The trial was \textit{`stopped early for safety reasons.'} \textbf{These findings show that no dexamethasone dose decreased the risk of postoperative bleeding compared to placebo: the 0.05 mg/kg dose had a higher rate, the 0.15 mg/kg dose was similar, and the 0.5 mg/kg dose had a much higher rate, with the difference being statistically significant.}

\end{example}
\subsection{PrivacyQA}
\begin{example}{Non-gold but thematically aligned sentences retrieved for a UAED question (gold sentences in bold).}
\textbf{Question:} Can i delete my search history?
\\

\textbf{Selected Sentences:}
    
\textit{0016}: In normal browsing mode, the Cake browser may store your browsing history (URLs of pages that you visit, your IP address, a cache of text, images and other resources from those pages) on your local device, in an anonymous fashion that is not tied to your identity.

\textbf{\textit{0020}: When you are in Privacy Mode, Cake does not collect your anonymous browsing history information and deletes any cookies or other downloaded information when you close any tabs that were opened in Privacy Mode.}

\textit{0022}: You can manage your information in the Cake browser.

\textbf{\textit{0023}: At any time, you can delete your browsing history (URLs of pages that you visit, a cache of text, images and other resources from those pages) and remove any cookies or other site data stored on your device.}
\end{example}

\begin{example}{Thematic drift into UCC sentences for a TPSC question (gold sentences in bold).}
\textbf{Question:} Can companies buy my info?
\\

\textbf{Selected Sentences:}
    
\textbf{\textit{0032}: Personalized Ads and Third-party Partners: We use the information we collect from all of our services to provide, maintain, protect and improve them, to develop new ones and to protect ourselves and our users.}

\textbf{\textit{0033}: We also use this information to offer you personalized ads.}

\textbf{\textit{0034}: We also have relevant third-party partners who offer you personalized ads.}

\textit{0035}: You can click here to know all such third-party partners name and their respective privacy policy and partner list (if any).

\textbf{\textit{0036}: These third-party partners will collect, use, store, share or other process (collectively Process) your personal information solely for the purpose of providing you personalized ads.}

\textit{0037}: If you dont agree on such third-party partners privacy policy or are unwilling them to Process your personal information, please do not agree or accept this TouchPal Privacy Policy.

\textit{0038}: If you want to stop receiving personalized ads, you can go to Settings of TouchPal Keyboard, once you chose not to receive personalized ads, we and our third-party partners will stop sending personalized ads to you.

\textit{0040}: Data that we do not collect: We will not collect or record your pin code entered in a password field, or your credit card entered in a secure text filed for whatever reason.

\textit{0041}: We will ask for your consent before using information for a purpose other than those set out in this Privacy Policy.

\textit{0047}: Disclosure Except as otherwise specified herein, we will not disclose your personal information or any other data that we collected or recorded to any third party unless you have consented to such disclosure or where we are required to do so by law.
\end{example}

\begin{example}{Relevant answer provided for a gold unanswerable question.}
\textbf{Question:} Is my information safe and secure from hackers?
\\

\textbf{Selected Sentences:} 

\textit{0020}: Remember that email may not necessarily be secure against interception.

\textit{0021}: If your email communication is very sensitive, or includes information such as your bank account, charge card, or social security number, you should send it by mail unless the Web site indicates the particular site is secured.

\textit{0110}: SecurityIn those instances where we secure your personal information in transit to us and upon receipt, the Postal Service uses the industry standard encryption software, Secured Socket Layer (SSL).

\textit{0111}: The URL in your browser will change to HTTPS instead of HTTP when this security feature is invoked.

\textit{0113}: For site security purposes and to ensure that this service remains available to all users, the Postal Service employs software programs to monitor network traffic in order to identify unauthorized attempts to upload or change information, or otherwise cause damage.

\textit{0114}: Except for authorized law enforcement investigations, no other attempts are made to identify individual users or their usage habits other than those uses identified in this policy.

\textit{0115}: Unauthorized modification or misuse of information stored in this system will be investigated and may result in criminal prosecution.

\textit{0117}: Postal Service information systems may be protected by EINSTEIN cybersecurity capabilities, under the operational control of the U.S. Department of Homeland Security's United States Computer Emergency Readiness Team (US-CERT).

\textit{0118}: Electronic communications with the Postal Service may be scanned by government-owned or contractor equipment to look for network traffic indicating known or suspected malicious cyber activity, including malicious content or communications.

\textit{0119}: Electronic communications within the Postal Service will be collected or retained by US-CERT only if they are associated with known or suspected cyber threats.

\textit{0120}: US-CERT will use the information collected through EINSTEIN to analyze the known or suspected cyber threat and help the Postal Service and other agencies respond and better protect their computers and networks.

\textit{0124}: The USPS holds service providers to the same rigorous standards of privacy that apply to the USPS, including being made subject to the Privacy Act and USPS privacy policies.
\end{example}

\begin{example}{Poor answer retrieved for a gold unanswerable question.}
\textbf{Question:} Do you store any personally identifiable data even after the app is removed?
\\

\textbf{Selected Sentences:}
       
\textit{0026}: You may access or modify your washingtonpost.com account on the User Profile page.

\textit{0053}: We also may collect other information about you, your device, or your use of the services in ways that we describe to you at the point of collection or otherwise with your consent.

\textit{0094}: We reserve the right to transfer any information we have about you in the event that we sell or transfer all or a portion of our business or assets to a third party, such as in the event of a merger, acquisition, or in connection with a bankruptcy reorganization.

\textit{0130}: You also may uninstall our apps using the standard uninstall process available on your mobile device or app marketplace.
\end{example}

\subsection{e-SNLI}
\begin{example}{Corrected label after gap-aware rationale refinement (highlighted text = recognized gold reason).}
\textbf{Context}: A boy and a dog run through the grass.

\textbf{Statement}: A boy and his dog run through field.

\textbf{Gold Label}: \textit{neutral}

\textbf{Gold Reason}: Just because a boy and a dog run through the grass doesn't mean that he is running with his dog through the field.
\\

\textbf{Model Baseline Label:} \textit{entailment}$\times$

\textbf{Model Baseline Reason}: The context and statement both describe a boy and a dog running through grassy terrain, with `field' and `grass' being nearly synonymous in this context. The only minor difference is the use of `his dog' instead of `a dog,' but \textit{it is reasonable to infer possession.}
\\

\textbf{Model Final Label}: \textit{neutral}$\checkmark$

\textbf{Iter 2 Reference Resolution Explanation}: \hl{The output notes that `his dog' infers ownership not strictly stated,} but could be more explicit that the context only mentions `a dog' and does not specify any relationship, making the inference plausible but not certain. 

\textbf{Model Final Reason}: The statement is plausible given the context, \textit{but it assumes the dog belongs to the boy} and generalizes `grass' to `field,' both of \textit{which are not strictly stated} and introduce minor inferential gaps.

\end{example}

\begin{example}{Model label conflicting with gold but with detailed gap-based reasoning (highlighted text = recognized gold reason; italicized text = counter argument).}
\textbf{Context}: The man works hard in his home office.

\textbf{Statement}: Home based worker works harder.

\textbf{Gold Label}: \textit{entailment}

\textbf{Gold Reason}: Home office means he is a home based worker.
\\

\textbf{Model Baseline Label:} \textit{neutral}$\times$

\textbf{Model Baseline Reason}: The context states that a man works hard in his home office, but it does not compare his work ethic to others or claim that home-based workers work harder in general.
\\

\textbf{Model Final Label}: \textit{contradiction}$\times$

\textbf{Iter 1 Reference Resolution Explanation}: The output accurately matches `the man' in the context to `home based worker' in the statement, with no ambiguity or confusion.

\textbf{Iter 1 Lexical Inferences Explanation}: \hl{The output correctly interprets `home office' as indicating a `home based worker'} and `works hard' as related to `works harder,' but could more explicitly state that `hard' and `harder' are not equivalent.

\textbf{Iter 1 Pragmatic Inferences Explanation}: \textit{The output recognizes that the context only supports the man working hard, not that he works harder than someone else.} It could be slightly improved by noting that, pragmatically, `works harder' implies a comparison that is not inferable from the context.

\textbf{Model Final Reason}: The context says the man works hard, but does not mention anyone or anything to compare to, so the comparative claim that he works harder is unsupported; `works hard' does not entail `works harder.'

\end{example}

\begin{example}{Incorrect label flip accompanied by plausible multi-gap explanation.}
\textbf{Context}: Racers run for the finish line, in an indoor meet, and one man has a big lead.

\textbf{Statement}: Racers are running a marathon.

\textbf{Gold Label}: \textit{neutral}

\textbf{Gold Reason}: The racers are not necessarily participating in a marathon..
\\

\textbf{Model Baseline Label:} \textit{neutral}$\checkmark$

\textbf{Model Baseline Reason}: The context mentions an indoor meet, which typically involves shorter races, not marathons. Therefore, it is unlikely that the racers are running a marathon.
\\

\textbf{Model Final Label}: \textit{contradiction}$\times$

\textbf{Iter 1 Quantitative and Comparative Reasoning Explanation}: The output correctly notes that indoor meets typically involve shorter races, not marathons, which are 42.195 km and almost never run indoors. However, it could be more explicit about the quantitative mismatch between `marathon' and typical indoor race distances.

\textbf{Iter 1 Pragmatic Inferences Explanation}: The output uses commonsense knowledge about indoor meets and marathons, but could be firmer in stating that marathons are essentially never held indoors, making the statement nearly a contradiction.

\textbf{Model Final Reason}: Marathons are 42.195 km and never held indoors; indoor meets only have short races, so the statement directly contradicts the context.

\end{example}

\begin{example}{Complex multi-line reasoning explicitly surfaced by GIER (highlighted text = recognized gold reason; italicized text = key inferential shift between baseline and final reasoning).}
\textbf{Context}: At a men's basketball game, number 55, in white and yellow, tries desperately to prevent number ten, in black and gray, from making a shot while the rest of the players watch frozen on the court.

\textbf{Statement}: Number 55 is guarding number 10 one on one.

\textbf{Gold Label}: \textit{entailment}

\textbf{Gold Reason}: Preventing someone from making a shot is part of guarding them. The rest of the players watch frozen implies that number 55 is the only one guarding number 10.
\\

\textbf{Model Baseline Label:} \textit{entailment}$\checkmark$

\textbf{Model Baseline Reason}: The context describes number 55 trying to prevent number 10 from making a shot, which strongly suggests one-on-one guarding, \textit{though it does not explicitly rule out the presence of other defenders.}
\\

\textbf{Model Final Label}: \textit{entailment}$\checkmark$

\textbf{Iter 1 Logical Inferecens Explanation}: The output recognizes that the context strongly suggests one-on-one guarding but does not guarantee exclusivity, correctly identifying a small logical gap. However, \hl{it could more explicitly note that `the rest of the players watch frozen' further supports the one-on-one scenario.}

\textbf{Iter 1 Lexical Inferences Explanation}: \hl{The output correctly interprets `tries desperately to prevent' as guarding} and 'watch frozen' as not participating, but could be more explicit in connecting these phrases to the concept of one-on-one defense.

\textbf{Model Final Reason}: The context makes it clear that number 55 is guarding number 10 one on one, as the \textit{rest of the players are described as watching frozen and not participating, which excludes any other defenders.}

\end{example}

\begin{figure*}[t]

  \begin{tcolorbox}[title=SciFact GIER Initial Prompt]
\tiny

\textbf{{\small Task}}

Support or refute a claim.
\\

\textbf{{\small Input}}

\textbf{Text:}

\textsf{[Abstract]}
\\

\textbf{Claim:}

\textsf{[Claim]}
\\

\textbf{{\small Instruction}}

\textbf{SUPPORT} or \textbf{REFUTE} the claim based on the provided text. Then, write a rationale to justify the decision, while carefully avoid the following potential gaps.
\\

\textsf{[Enumerated List of Gap Definitions]}
\\

\textbf{\small Text Quoting Format}

-- When quoting sentences, segments, or phrases from the provided text in your rationale, always place the quoted text within \textbf{single quotes} (e.g., \verb +'quoted text'+).

-- If you abbreviate or omit parts of the quoted text, use \textbf{three ellipsis dots (...)} to indicate the omission. 

-- ! \textbf{Do not combine text from multiple sentences into a single quote}, even with ellipses. Each quote must be sourced from a \textbf{single sentence} only.
\\

\textbf{\small Output Format}

\textbf{Important:} Any deviation from this format (e.g., missing or extra commas, unquoted text, etc.) will make the output invalid. Only provide the valid JSON response, with no additional explanations or comments. 

Output your response in \textbf{JSON format} using the following template. 

\begin{verbatim}
```json
{
  "decision": "...",
  "rationale": "..."
}
\end{verbatim}

  \end{tcolorbox}
\caption{GIER initial prompt for SciFact rationale generation task.}
\label{fig:scifact_initial_prompt}

\end{figure*}

\begin{figure*}[t]

  \begin{tcolorbox}[title=PrivacyQA GIER Initial Prompt]
\tiny 

\textbf{\small Task}

Extractive sentence selection.
\\

\textbf{\small Input}

\textbf{Privacy Policy Text:}

Below is a JSON object containing a list of items. Each item is a sentence from a privacy policy and contains an \verb+id+ (the identifier) and the \verb+text+ of the sentence. 
\\

\textsf{[Privacy Policy Sentences]}
\\

\textbf{User Question:}

\textsf{[Question]}
\\

\textbf{\small Instruction}

1. \textbf{Answerability:} Determine if the question is answerable based on the content given in the privacy policy text.

\quad    -- If the question is answerable, return \verb+"answerable": true+.

\quad    -- If the question is not answerable, return \verb+"answerable": false+.

2. \textbf{Sentence Selection:}

\quad    -- If the question is answerable, identify the subset of sentences that contains the contextually relevant information required to answer the question. 

\qquad        -- Return only the \verb+id+ values of those sentences in a JSON list.

\qquad    -- If the question is unanswerable, return an empty list \verb+[]+.
    
3. \textbf{Gap Avoidance:} While performing the steps above, carefully avoid the following potential gaps:
\\

\quad \textsf{[Enumerated List of Gap Definitions]} 
\\

\textbf{\small Output Format}

\textbf{Important:} Any deviation from this format (e.g., missing or extra commas, unquoted text, etc.) will make the output invalid. Only provide the valid JSON response, with no additional explanations or comments.

Output your response in \textbf{JSON format} using the following template. 

\begin{verbatim}
```json
{
  "answerable": true or false,
  "selected_sentence_ids": ["id_0001", "id_0002", ...]
}
\end{verbatim}
  \end{tcolorbox}
\caption{GIER initial prompt for PrivacyQA sentence selection task.}
\label{fig:privacyqa_initial_prompt}

\end{figure*}

\begin{figure*}[t]

  \begin{tcolorbox}[title=e-SNLI GIER Initial Prompt]
\tiny 

\textbf{\small Task}

Natural language inference.
\\

\textbf{\small Input}

\textbf{Context:}

\textsf{[Context]}
\\

\textbf{Statement:}

\textsf{[Statement]}
\\

\textbf{\small Instructions}
    
1. \textbf{Entailment Score}: Assign an entailment score between 0 and 10 to indicate how likely is that the statement is an entailment from the context. A score of 0 implies not at all likely (a contradiction) and a score of 10 implies certainty. 

2. \textbf{Reason}: Output the reason for the score in one or two sentences.
    
3. \textbf{Gap Avoidance}: While performing the steps above, carefully avoid the following potential gaps:
\\

\quad\textsf{[Enumerated List of Gap Definitions]} 
\\

\textbf{\small Output Format}

\textbf{Important}: Any deviation from this format (e.g., missing or extra commas, unquoted text, etc.) will make the output invalid. Only provide the valid JSON response, with no additional explanations or comments.

Output your response in \textbf{JSON format} using the following template. 

\begin{verbatim}
```json
{
  "entailment_score": X,
  "reason": "..."
}
\end{verbatim}

  \end{tcolorbox}
\caption{GIER initial prompt for e-SNLI commonsense inference task.}
\label{fig:esnli_initial_prompt}

\end{figure*}

\begin{figure*}[t]

  \begin{tcolorbox}[title=SciFact GIER Revision Prompt]
\tiny

\textbf{\small Context}

You earlier wrote a rationale for why a claim is supported or refuted by a given piece of text. 
\\

\textbf{\small Task}

Perform gap analysis and revise decision and/or rationale.
\\

\textbf{\small Input}

\textbf{Text:}

\textsf{[Abstract]}
\\

\textbf{Claim:}

\textsf{[Claim]}
\\

\textbf{Your Previous Gap Analyses:}

Here are the rationales, explanations, and scores of your previous revisions in JSON format. Review and consider this history during revision.

\textsf{[All Previous Responses]}
\\

\textbf{Your Most Recent Decision:}

\textsf{[Most Recent Decision]}
\\

\textbf{Your Most Recent Rationale:}

\textsf{[Most Recent Rationale]}
\\

\textbf{\small Gaps}

Carefully consider the following \textsf{[Gap Count]} gap definitions. These gaps may or may not be present in your rationale. You will be asked to use them in a subsequent instruction.
\\

\textsf{[Enumerated List of Gap Definitions]}
\\

\textbf{\small Instructions}

\textbf{Important:} Each gap is equally important and should be addressed with the same level of attention. Do not prioritize one gap over another. Carefully consider each gap when performing your analysis and revision.

Follow these steps:

1.  \textbf{Gap Analysis Of Most Recent Rationale:} For each of the \textsf{[Gap Count]} gaps listed above:

\quad -- Assign a score between 0 and 10 using the following scale:

\qquad         -- 0: Your rationale entirely fails to address the gap. No attempt is made to meet the criterion.

\qquad         -- 1--3: Your rationale shows minimal effort to close the gap. The issue is prominent, with several major problems remaining.

\qquad         -- 4--6: Your rationale attempts to address the gap, but inconsistently or insufficiently. Significant room for improvement remains.

\qquad         -- 7--9: Your rationale mostly addresses the gap, with only minor or occasional lapses that could be refined.

\qquad         -- 10: Your rationale has absolutely no room for improvement with respect to the gap. 

\quad-- Write a short explanation evaluating your rationale based on the gap. If applicable, identify specific improvements that could be made.

2.  \textbf{Compare Current Analysis With Previous:} Review the rationales, scores, and explanations in your current and previous analyses. For each gap:

\quad    -- Identify areas where your most recent rationale improves, regresses, or stagnates compared to earlier attempts.

\quad    -- If a previous rationale handled something better, try to retain or reintroduce that improvement.

\quad    -- Avoid repeating flaws that were previously identified and addressed.
    
3. \textbf{Consolidate Across Iterations:} Write a consolidated explanation of how you intend to integrate insights from previous iterations and your current analysis.

4.  \textbf{Revise Decision and Rationale:} Based on the insights from Steps 1, 2, and 3, revise your decision and rewrite your rationale to justify the decision, while improving it with respect to the gaps, if necessary.
\\

\textbf{\small Text Quoting Format}

\textit{<same as in initial prompt>}
\\

\textbf{\small Output Format}

\textbf{Important}: Any deviation from this format (e.g., missing or extra commas, unquoted text, etc.) will make the output invalid. Only provide the valid JSON response, with no additional explanations or comments. 

Return your output in \textbf{JSON format} using the following template. 
\begin{verbatim}
```json
{
  "gap_analysis": {
    "Coverage": {"score": X, "explanation": "..." },
    "Conciseness": {"score": X, "explanation": "..." },
    ...
  },
  "consolidated_explanation": "...",
  "revised_decision": "...",
  "revised_rationale": "..."
}
\end{verbatim}
  \end{tcolorbox}
\caption{GIER revision prompt for SciFact rationale generation task.}
\label{fig:scifact_revision_prompt}

\end{figure*}

\begin{figure*}[t]

  \begin{tcolorbox}[title=PrivacyQA GIER Revision Prompt]
\tiny 
\textbf{\small Context}

You previously determined if a question is answerable based on a privacy policy, and selected a group of relevant sentences when the question was answerable.
\\

\textbf{\small Task}

Perform gap analysis and revise decision and/or selection.
\\

\textbf{\small Input}

\textbf{Privacy Policy Text:}

Below is a JSON object containing a list of items. Each item is a sentence from the privacy policy and contains an \verb+id+ (the identifier) and the \verb+text+ of the sentence. 
\\

\textsf{[Privacy Policy Sentences]}
\\

\textbf{User Question:}

\textsf{[Question]}
\\

\textbf{Your Previous Gap Analyses:}

Here are the answerability decisions, selected sentence identifiers, and scores of your previous revisions in JSON format. Review and consider this history during revision.
\\

\textsf{[All Previous Responses]}
\\

\textbf{Your Most Recent Answerability Decision:}

\textsf{[Most Recent Decision]}
\\

\textbf{Your Most Recent Sentence Selections:}

\textsf{[Most Recent Selections]}
\\

\textbf{\small Gaps}

Carefully consider the following \textsf{[Gap Count]} gap definitions. These gaps may or may not be present in your analysis. You will be asked to use them in a subsequent instruction.
\\

\textsf{[Enumerated List of Gap Definitions]}
\\

\textbf{\small Instructions}

\textbf{Important:} Each gap is equally important and should be addressed with the same level of attention. Do not prioritize one gap over another. Carefully consider each gap when performing your analysis and revision.

Follow these steps:

1.  \textbf{Gap Analysis Of Most Recent Selections:} For each of the \textsf{[Gap Count]} gaps listed above:

\quad    -- Assign a score between 0 and 10 using the following scale:

\qquad        -- 0: Your selections entirely fail to address the gap. No attempt is made to meet the criterion.

\qquad        -- 1--3: Your selections show minimal effort to close the gap. The issue is prominent, with several major problems remaining.

\qquad        -- 4--6: Your selections attempt to address the gap, but inconsistently or insufficiently. Significant room for improvement remains.

\qquad        -- 7--9: Your selections mostly address the gap, with only minor or occasional lapses that could be refined.

\qquad        -- 10: Your selections have absolutely no room for improvement with respect to the gap. 

\quad    -- Write a short explanation evaluating your selections based on the gap. If applicable, identify specific improvements that could be made.

2.  \textbf{Compare Current Analysis With Previous}: Review the decisions, selections, and explanations in your current and previous analyses. For each gap:

\quad    -- Identify areas where your most recent decision and selections improve, regress, or stagnate compared to earlier attempts.

\quad    -- If a previous decision and selections handled something better, try to retain or reintroduce that improvement.

\quad    -- Avoid repeating flaws that were previously identified and addressed.
    
3. \textbf{Consolidate Across Iterations}: Write a consolidated explanation of how you intend to integrate insights from previous iterations and your current analysis.

4.  \textbf{Revise Decision and Selection}: Based on the insights from Steps 1, 2, and 3, revise your answerability decision (\verb+true+ or \verb+false+) and the list of selected sentence \verb+id+s, improving the output with respect to the gaps, if necessary.
\\

\textbf{\small Output Format}

\textbf{Important}: Any deviation from this format (e.g., missing or extra commas, unquoted text, etc.) will make the output invalid. Only provide the valid JSON response, with no additional explanations or comments. 

Return your output in \textbf{JSON format} using the following template. 

\begin{verbatim}    
```json
{
  "gap_analysis": {
    "Coverage": {"score": X, "explanation": "..." },
    "Thematic Overreach": {"score": X, "explanation": "..." }
  },
  "consolidated_explanation": "...",
  "revised_answerable": true or false,
  "revised_selected_sentence_ids": ["id_0001", "id_0002", ...]
}
\end{verbatim}

  \end{tcolorbox}
\caption{GIER revision prompt for PrivacyQA sentence selection task.}
\label{fig:privacyqa_revision_prompt}

\end{figure*}

\begin{figure*}[t]

  \begin{tcolorbox}[title=e-SNLI GIER Revision Prompt]
\tiny 
\textbf{\small Context}

You previously provided an entailment score for a statement with respect to a provided context, and provided a reason for the score.
\\

An entailment score is a number between 0 and 10 to indicate how likely is that a statement is an entailment from a context. A score of 0 implies not at all likely (a contradiction) and a score of 10 implies certainty. 
\\

\textbf{\small Task}

Perform gap analysis and revise entailment score and/or reason.
\\

\textbf{\small Input}

\textbf{Context:}

\textsf{[Context]}
\\

\textbf{Statement:}

\textsf{[Statement]}
\\

\textbf{Your Previous Gap Analyses:}

Here are the entailment scores, reasons, and gap scores of your previous revisions in JSON format. Review and consider this history during revision.
\\

\textsf{[All Previous Responses]}
\\

\textbf{Your Most Recent Entailment Score:}

\textsf{[Most Recent Entailment Score]}
\\

\textbf{Your Most Recent Reason:}

\textsf{[Most Recent Reason]}
\\

\textbf{\small Gaps}

Carefully consider the following \textsf{[Gap Count]} gap definitions. These gaps may or may not be present in your analysis. You will be asked to use them in a subsequent instruction.
\\

\textsf{[Enumerated List of Gap Definitions]}
\\

\textbf{\small Instructions}

\textbf{Important}: Each gap is equally important and should be addressed with the same level of attention. Do not prioritize one gap over another. Carefully consider each gap when performing your analysis and revision.
    
Follow these steps:

1.  \textbf{Gap Analysis Of Most Recent Output:} For each of the \textsf{[Gap Count]} gaps listed above:

\quad    -- Assign a score between 0 and 10 using the following scale:

\qquad        -- 0: Your output entirely fails to address the gap. No attempt is made to meet the criterion.

\qquad        -- 1--3: Your output shows minimal effort to close the gap. The issue is prominent, with several major problems remaining.

\qquad        -- 4--6: Your output attempts to address the gap, but inconsistently or insufficiently. Significant room for improvement remains.

\qquad        -- 7--9: Your output mostly addresses the gap, with only minor or occasional lapses that could be refined.

\qquad        -- 10: Your output has absolutely no room for improvement with respect to the gap. 

\quad    -- Write a short explanation evaluating your output based on the gap. If applicable, identify specific improvements that could be made.

2.  \textbf{Compare Current Analysis With Previous:} Review the entailment scores, reasons, and explanations in your current and previous analyses. For each gap:

\quad    -- Identify areas where your most recent output improves, regresses, or stagnates compared to earlier attempts.

\quad    -- If a previous output handled something better, try to retain or reintroduce that improvement.

\quad    -- Avoid repeating flaws that were previously identified and addressed.
    
3. \textbf{Consolidate Across Iterations:} Write a consolidated explanation of how you intend to integrate insights from previous iterations and your current analysis.

4.  \textbf{Revise Entailment Score and Reason:} Based on the insights from Steps 1, 2, and 3, revise your entailment score and the reason, improving the output with respect to the gaps, if necessary. The reason should be brief, within one or two sentences.
\\

\textbf{\small Output Format}

\textbf{Important}: Any deviation from this format (e.g., missing or extra commas, unquoted text, etc.) will make the output invalid. Only provide the valid JSON response, with no additional explanations or comments. 

Return your output in \textbf{JSON format} using the following template. 

\begin{verbatim}
```json
{
  "gap_analysis": {
    "Quantitative and Comparative Reasoning": {"score": X, "explanation": "..." },
    "Reference Resolution": {"score": X, "explanation": "..." },
    ...
  },
  "consolidated_explanation": "...",
  "revised_entailment_score": X,
  "revised_reason": "..."
}
\end{verbatim}
  \end{tcolorbox}
\caption{GIER revision prompt for e-SNLI commonsense inference task.}
\label{fig:esnli_revision_prompt}

\end{figure*}

\end{document}